\documentclass[sn-mathphys,Numbered]{sn-jnl}

\usepackage{graphicx}%
\usepackage{multirow}%
\usepackage{amsmath,amssymb,amsfonts}%
\usepackage{amsthm}%
\usepackage{mathrsfs}%
\usepackage[title]{appendix}%
\usepackage{xcolor}%
\usepackage{textcomp}%
\usepackage{manyfoot}%
\usepackage{booktabs}%
\usepackage{algpseudocode}%
\usepackage{listings}%
\usepackage[inline]{enumitem}
\usepackage{url}
\usepackage{breakurl}
\usepackage{hyperref}

\usepackage{amssymb}
\usepackage[T1]{fontenc}
\usepackage{comment}
\usepackage{subcaption}

\usepackage{multirow}
\usepackage{amssymb}
\usepackage{pifont}
\usepackage{soul}
\usepackage{tabularx}
\usepackage{booktabs}
\usepackage{todonotes}
\usepackage{makecell}

\usepackage{listings,chngcntr}
\usepackage[linesnumbered]{algorithm2e} 
\AtBeginEnvironment{algorithm}{}

\SetCommentSty{mycommfont}

\begin{document}

\title[Mining Frequent Structures in Conceptual Models]{Mining Frequent Structures in Conceptual Models}


\author[1]{Mattia Fumagalli}
\author[2]{Tiago Prince Sales}
\author[2]{Pedro Paulo F. Barcelos}
\author[3]{Giovanni Micale}
\author[5]{Philipp-Lorenz Glaser}
\author[5]{Dominik Bork}
\author[4]{Vadim Zaytsev}
\author[1]{Diego Calvanese}
\author[2]{Giancarlo Guizzardi}

\affil[1]{\footnotesize{KRDB Research Centre on Knowledge and Data, Free University of Bozen-Bolzano,
            Bolzano,
            Italy}}

\affil[2]{\footnotesize{Semantics, Cybersecurity \& Services (SCS), University of Twente,
            Enschede,
            The Netherlands}}

\affil[3]{\footnotesize{Department of Clinical and Experimental Medicine, University of Catania,
    Catania,
    Italy}}

\affil[4]{\footnotesize{Formal Methods and Tools (FMT), University of Twente,
            Enschede,
            The Netherlands}}

\affil[5]{\footnotesize{TU Wien, Business Informatics Group, Vienna, Austria}\vspace{-2em}}

\abstract{The challenge of using structured methods to represent knowledge is a well-documented issue in conceptual modeling and has been the focus of extensive research. It is widely recognized that adopting \textit{modeling patterns} offers an effective structural approach for designing conceptual models. Patterns, in this context, refer to generalizable, recurring structures that provide solutions to common design problems. They significantly enhance both the \textit{understanding} and \textit{improvement} of the modeling process. Numerous experimental studies have demonstrated the undeniable value of using patterns in conceptual modeling. Despite this, the task of identifying patterns in conceptual models remains highly complex, and there is currently no systematic method for pattern discovery. To address this gap, this paper proposes a general approach for \textit{discovering frequent structures} in conceptual modeling languages as a means to support pattern identification. Specifically, we focus on uncovering recurring structures that reflect the usage patterns of a given conceptual modeling language. As proof of concept, we implement our approach by focusing on two widely-used conceptual modeling languages. This implementation includes an exploratory tool that integrates a \textit{frequent subgraph mining algorithm} with \textit{graph manipulation techniques}. The tool processes multiple conceptual models and identifies recurrent structures based on various criteria. We validate the tool using two state-of-the-art curated datasets: one consisting of models encoded in OntoUML and the other in ArchiMate. The primary objective of our approach is to provide a support tool for language engineers. This tool can be used to identify both effective and ineffective modeling practices, enabling the refinement and evolution of conceptual modeling languages. Furthermore, it facilitates the reuse of accumulated expertise, ultimately supporting the creation of higher-quality models in a given language.}




\maketitle

\section{Introduction}
\label{s:intro}


Conceptual modeling is a highly complex task. As empirical results show~\cite{guizzardi2010theoretical,gurlebeck2020conceptual,van1997validation,kayama2014practical,reder2010model}, this is often due to multiple factors, such as the cognitive limitation of modelers, the lack of information about the domain to be modeled, and the nature of the conceptual modeling language being adopted. For this reason, one of the main concerns of language engineers is overcoming modelers' difficulties and devising conceptual modeling languages that assure as much as possible the high-level quality of the output conceptual model. Patterns play a key role in this regard. These recurrent structures are \textit{generalizable} solutions to design problems that help in \textit{understanding} and \textit{improving} the process of creating models. Specifically, patterns represent examples of good (or bad) recurring practices that language engineers can identify by looking at the practical application of the language itself. By discovering and adopting recurrent structures, language engineers can then evolve the modeling language itself, speeding up and facilitating its application and easing the reuse of working experiences, by also helping to avoid common errors or misconceptions.

During the past decade, modeling patterns have been widely adopted by language designers~\cite{falbo2013ontology,guizzardi2014ontological} and, consequently, discovering them from design experiences has become of paramount importance. Still, the task of discovering recurring modeling structures across conceptual models presents multiple challenges. For instance, consider that 
\begin{enumerate*}[label=\textit{.\roman*}]
\item recurrent modeling practices can only be discovered by observing a vast number of conceptual modeling examples, and 
\item the assessment of the conceptual models and the identification of recurrent modeling structures often require analysis activities that are highly time-consuming when performed manually, as seen in processes like \textit{modularization}, \textit{frequency calculation}, or \textit{constructs correlation}.
\end{enumerate*}

The analysis of conceptual patterns and the discovery of recurring structures or useful patterns has emerged as a distinct research path that has seen significant growth in recent years. This research, as evidenced by seminal works like~\cite{gangemi2009ontology,falbo2013ontology,guizzardi2014ontological,besheli2018pattern}, focuses on analyzing reference patterns—used as case studies—to infer modeling strategies for specific problems. More recent work has explored the application of automated techniques to support heuristic tasks traditionally performed manually~\cite{fumagalli2022pattern,lawrynowicz2018discovery,skouradaki2016rose,mitra2017discovering}.

In this article, we join the ongoing research efforts to develop automated support for facilitating the empirical discovery of recurring modeling structures. Our contribution is an approach that integrates several state-of-the-art techniques, designed as an exploratory tool to enable users to interactively discover and analyze frequent structures. The implementation of this approach is based on gSpan~\cite{yan2002gspan}, a widely recognized \textit{frequent subgraph mining algorithm}. Through a command-line interface, this tool enables users to:

\begin{enumerate}[label=\textit{.\roman*}]
\item select the models to be processed, 
\item prepare the data to be mined by focusing on specific information and filtering out concepts that might not be relevant for that type of analysis, 
\item select certain features that the output frequent structures should have (e.g., number of nodes, frequency, or dissimilarities from known patterns), 
\item visualize the output frequent structures and query the input models to verify the domain occurrences of that structure, and
\item cluster the output structures to simplify the user's final assessment. 
\end{enumerate}

The main scope of our solution is to offer a support facility for language engineers that aims at exploiting bad/good practices to evolve and maintain the conceptual modeling language but also to favor the reuse of encoded experience in designing better models with the given language.

We evaluate our approach using two large and curated state-of-the-art datasets of frequently used conceptual modeling languages, in particular, the OntoUML dataset~\cite{sales2023fair} and the ArchiMate dataset~\cite{GlaserSB23}. We adopted datasets encoded in OntoUML and ArchiMate languages for several reasons. Primarily, these languages are based on pre-defined patterns, which are frequently used across different models to address common modeling issues. This allows us to evaluate our approach to finding useful structures that are already known by the designers of each language. Second, for our evaluations, we can access curated high-quality datasets, encoded in a uniform format, thus improving our findings' comprehensibility and reliability. Third, for our evaluations and experiment design, we can obtain feedback directly from the authors of the languages,\footnote{In particular, we leveraged this feedback in the OntoUML scenario.} showing how this may help enabling us to gather valuable information about the practical utility of the approach. Finally, OntoUML and ArchiMate have a variety of constructs and features that allow us to assume that if our approach works for these languages, it also works for widely used standard languages, such as UML or BPMN.

The outline of this paper is as follows: \autoref{s:baseline} explores the concept of ``pattern'' in conceptual modeling, focusing particularly on ``recurrent modeling structures''. This section also provides a brief overview of the primary technique employed for automating the mining process. \autoref{s:requirements} enumerates the requirements that guided the design of our methodology. In \autoref{s:method}, we introduce our framework, elaborating on its workflow and individual components. \autoref{s:implementation} details the implementation of our approach. In \autoref{s:eval} and \autoref{s:demo}, we discuss the experiments and demonstrations conducted to validate our solution. In \autoref{s:discussion}, we analyze insights gained from the evaluation and demonstration processes. \autoref{s:relatedwork} situates our work within the context of existing literature. Finally, \autoref{s:considerations} offers reflections on our findings.

\section{Research Baseline}
\label{s:baseline}

\subsection{Patterns in Conceptual Modeling}

Currently, our proposal primarily focuses on patterns in conceptual models. These model fragments represent recurrent structures formed by modeling constructs from a specific conceptual modeling language. Keeping track of these patterns allows one to understand how a language is applied, thus offering powerful means to language designers for maintaining and evolving the language itself. Using recurrent structures, one can, e.g., discover the most common combinations of language constructs, identify language dialects for specific application domains, verify possible restrictions in using constructs or their combinations, and determine the frequent \textit{subversions} of the language. The notion of \textit{systematic language subversion}~\cite{guizzardi2015towards} refers to an ungrammatical use of a language's constructs that becomes recurrent in a language community signalling a design limitation of that language. It is closely related to \textit{coding traditions}~\cite{Zen2021} that cover coding policies, notational guidelines, naming conventions, implementation patterns, programming idioms, etc.

\begin{figure}[ht]
\centering
\includegraphics[width=0.7\textwidth]{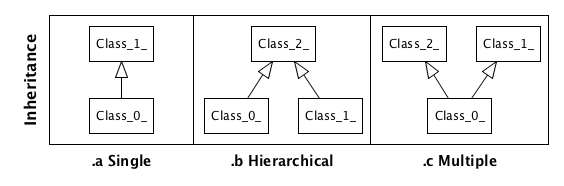}
\caption{Example of inheritance recurrent structures in UML class diagrams \cite{bruegge2009object}.}
\label{UMLpatterns}
\end{figure}

\autoref{UMLpatterns} shows examples of common usages of the generalization relationship in UML class diagrams. These structures represent the ways by which modelers represent intentional sub-typing (inheritance) between classes (see \textit{.a}, \textit{.b} and \textit{.c} types), and can be taken as useful insights for understanding the language application. For instance, given a set of models about a certain domain, it is possible to discover that the \textit{multiple inheritance} (\textit{.c}) is used to capture class compositionally~\cite{bruegge2009object}. This observation might trigger language designers to suggest class composition relations when modelers use two or more generalization relations for a class.

\begin{figure}[ht]
\centering
\includegraphics[width=0.9\textwidth]{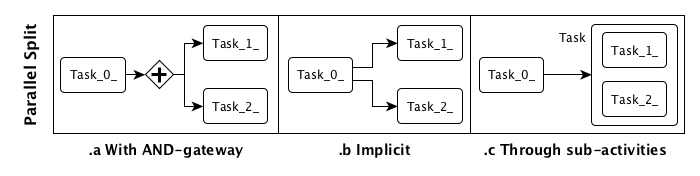}
\caption{Example of parallel split recurrent structures in BPMN diagrams \cite{wohed2006pattern}.}
\label{BPMNpatterns}
\end{figure}

Similarly, \autoref{BPMNpatterns} represents three examples of recurring structures in BPMN models. These patterns correspond to equivalent ways of modeling tasks parallel splitting (see \textit{.a}, \textit{.b}, and \textit{.c} types) and are related to how people design control flows. Just as with UML structural patterns, these BPMN fragments can offer interesting insights into the language's usage. For instance, by observing multiple occurrences of the implicit parallel split (.b), language designers may decide to force modelers to make explicit the gateway for the splitting, to avoid unintended instantiations of the model (i.e., where the splitting may be intended as an \texttt{``OR''}, instead of an \texttt{``AND''}, depending on the modelers' scope).

\subsection{Frequent Subgraph Mining}

Frequent Subgraph Mining (FSM) is a well-known technique~\cite{jiang2013survey} used to find frequent subgraphs in a graph dataset. This technique typically involves a dual-phase approach. The initial stage involves the \textit{creation of subgraph candidates}~\cite{jiang2013survey}, where subgraphs in the graph dataset are searched and proposed for the analysis, while the subsequent step entails \textit{evaluating the frequency} of the generated subgraphs to ascertain their prevalence. 

FSM presents two distinct problem formulations: 
\begin{enumerate*}[label=\textit{.\roman*}]
\item \textit{graph transaction-based FSM}, and 
\item \textit{single graph-based FSM}. 
\end{enumerate*}
In graph transaction-based FSM, the input dataset comprises a collection of medium-sized graphs termed ``transactions''~~\cite{agrawal1994fast}. On the other hand, single graph-based FSM uses a single, significantly large graph as input. In this setting, a subgraph, denoted as \textit{g}, is deemed frequent if its occurrence count surpasses a predefined threshold value, commonly referred to as the \textit{support threshold}. The support calculation varies according to the problem formulation. In a transactional dataset, the support $\alpha\:(0 < \alpha <1 )$ of a graph is the ratio of the number of transactions to which
this graph occurs to the total number of transactions. In a single large graph, the support of a graph is the number of its occurrences in this graph. 

Formally, given a graph $G\:=\:(V_g\:,\:E_g)$, where $V_g$ is a set of (possibly labeled) vertices and $E_g$ is a set of (possibly labeled) edges, then a graph $H = (V_h, E_h)$ is a sub-graph of G if and only if its vertices and edges are a subset of the vertices $(V_h\:\subseteq\:V_g)$ and edges $(E_h \:\subseteq\:E_g)$ of graph G, and the vertex subset include all endpoints of the edge subset. A subgraph is considered frequent when its support equals or exceeds the user-defined minimum support threshold.

\begin{figure}[h!]
\centering
\includegraphics[width=0.7\textwidth]{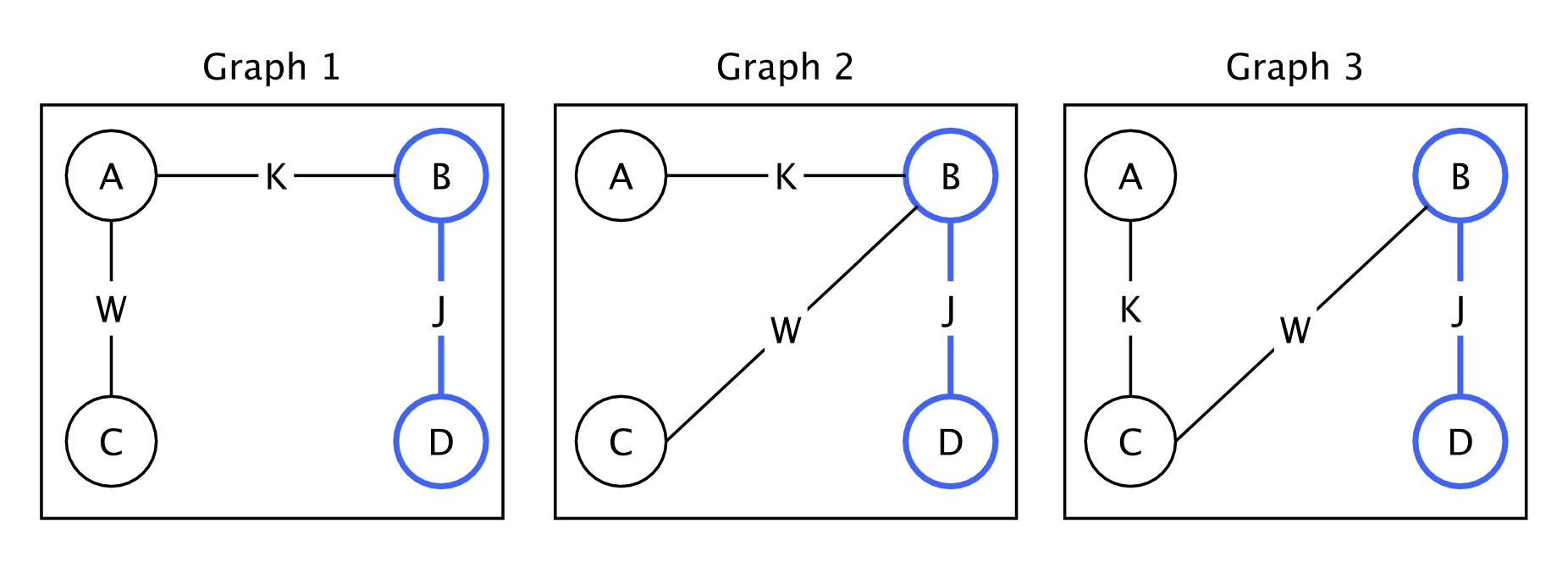}
\caption{Example of three graphs, with subgraph $B - J - D$ having \textit{support} \textit{(frequency)} of 3}
\label{graphdb}
\end{figure}

Let us consider the three graphs represented in \autoref{graphdb}. Suppose that we want to discover all the (connected) subgraphs with \textit{at least two nodes} that occur in \textit{at least three graphs}. By adopting \textit{minimum support 3} as a parameter and applying an FSM algorithm, we can obtain the set of all subgraphs appearing in at least three graphs, namely $B - J - D$.

In practical scenarios, different tools or algorithms can identify frequent subgraphs. In this paper, we adopt gSpan~\cite{yan2002gspan}, a popular state-of-the-art solution. This algorithm is often used for discovering frequent subgraphs in large graph databases and has variants that work on conceptual graphs~\cite{Faci2021cgSpan}, typed linked data~\cite{Zhang2012linked}, chemical compound data~\cite{Han2007FSP}, large disk-based databases~\cite{Wang2004disk}, data with differential privacy~\cite{Xing2021DP}, etc. It uses an approach based on a \textit{minimum DFS code}~\cite{tarjan1972depth} that allows the efficient generation of candidate subgraphs and pruning of infrequent ones. The algorithm iteratively discovers all frequent subgraphs containing an edge and then expands the search to include larger subgraphs. The database is shrunk as the search continues, and only graphs containing the current subgraph are considered. Studies show that gSpan can mine large frequent subgraphs with lower frequency and high performance~\cite{jiang2013survey}. 

\section{Requirements}
\label{s:requirements}

Our design for the proposed approach is based on an initial \textit{problem identification} activity. During this phase, we gathered feedback from five potential users of our approach: expert language engineers, who have been involved in the development of different conceptual modeling languages and who have been working on the identification of conceptual modeling patterns. We performed open-ended interviews and the main open questions we asked were: 

\begin{enumerate}[label=\textit{.\roman*}]
\item \textit{What is the relevance of an approach for facilitating the empirical discovery of structural modeling patterns in conceptual models?}, and 
\item \textit{What is required in order to facilitate the empirical discovery of modeling patterns?}
\end{enumerate}

This preliminary step helped us improve our awareness of the problem, better understanding the related work, and better identifying the features that our solution should offer to the end-users. The feedback from experts was crucial in defining both functional and non-functional requirements, which are needed to design the approach and evaluate the artifact in which our contribution is embedded. 

We mapped the key features that specify \textit{what} our approach should do into the following \textit{functional requirements}:
\vspace{0.5em}
\begin{itemize}
    \item[\textbf{R1.}] \textit{Interestingness:} The approach should facilitate the discovery of \textit{subjectively interesting} patterns, namely recurrent structures that can be considered interesting according to the user's interpretation. Here, the notion of ``subjectively interesting'' is inspired by the work from 
    Silberschatz and Tuzhilin~\cite{silberschatz1995subjective}, where a pattern is ranked as interesting by a user mainly because
    \begin{enumerate*}[label=\textit{.\roman*}]
    \item it is considered \textit{exploitable} for the modeling activities, or
    \item it contradicts some users' expectations.
    \end{enumerate*}
    \item[\textbf{R2.}] \textit{Customization:} The approach should allow engineers to manipulate the input models to obtain different patterns, which may vary in size (number of nodes or edges), may be related to different constructs (see, for instance, \textit{taxonomical} vs. \textit{non-taxonomical} structures), or maybe taken at different levels of granularity (e.g., in UML \textit{compositions} and \textit{aggregations} relations may be taken just as \textit{associations}). The approach should also allow the user to filter the output of the discovery process according to custom parameters. 
    \item[\textbf{R3.}] \textit{Comprehension:} The approach should assist engineers in the process of assessing and analyzing the output structures. This should be feasible by accounting for multiple types of frequency, e.g., \textit{relative frequency}, as the number of occurrences of a structure in each model; \textit{total frequency}, as the overall occurrences for a given pattern; \textit{frequency across models}, as the number of models in which a pattern occurs. Moreover, this should be supported by a human-readable visualization format of the output structures. 
\end{itemize}
\vspace{0.5em}
We mapped the key features that specify \textit{how} the approach should perform its functions into the following \textit{non-functional requirements}:

\vspace{0.5em}
\begin{itemize}
 \setlength{\itemindent}{1.0em}
    \item[\textbf{R4.}] \textit{Performance}. The approach should outperform the human discovery activity in terms of time. Moreover, the conceptual models processing and the mining step should \textit{happen} in a reasonable amount of time, even with a \textit{large} amount of data, where ``reasonable'' and ``large'' are considered concerning related work \cite{garcia2019big,mabroukeh2010taxonomy} (e.g., hundreds of models).
    \item[\textbf{R5.}] \textit{Compatibility}. 
    The approach should be able to support pattern discovery in multiple conceptual modeling languages.
\end{itemize}

\section{Method}
\label{s:method}

Our approach is represented as a workflow, consisting of several tasks, which can be categorized into three main phases: \textit{preparation}, \textit{discovery}, and \textit{assessment}. These phases allow the user to intervene in the workflow whose
inputs, outputs, and dependencies are combined as from \autoref{approach} below.

\begin{figure}[ht]
\centering
\includegraphics[width=1\textwidth]{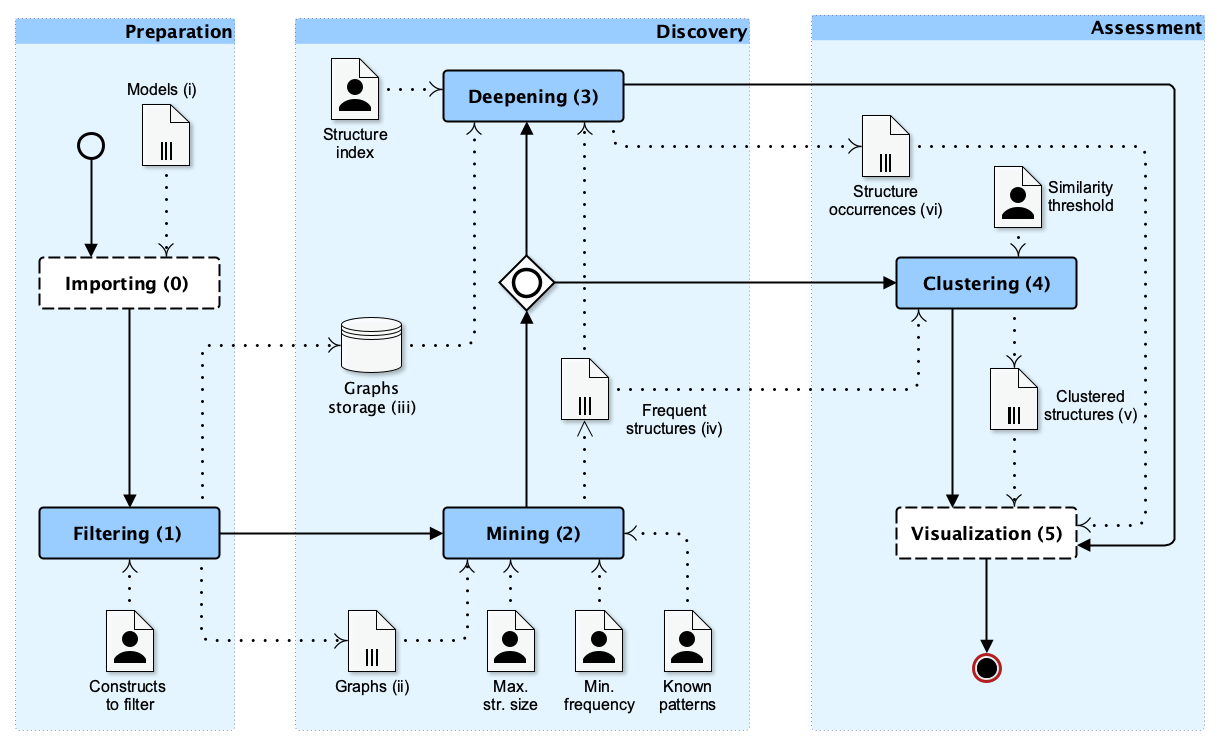}
\caption{The frequent structures discovery workflow.}
\label{approach}
\end{figure}

\subsection{Preparing the Input Data}
\label{s:preparation}

The ``Preparation'' phase focuses on transforming the input conceptual models in a format that is processable by the subsequent steps. In this phase, we have two main tasks. 
First, the \textit{Importing (0)} task consists of taking a set of conceptual models $M$ encoded in a language (e.g., UML or BPMN) and transforming each model $m_{i} \in M$ into a \textit{Labeled Property Graph (LPG)} $g_{j}$. Such a task is denoted by a white box in \autoref{approach} because it is language-dependent, namely it requires an \textit{ad hoc} transformation for each source conceptual modeling language taken into consideration. At first glance, this might seem straightforward since conceptual models are essentially graphs. That, however, is not always the case. Consider, for instance, the transformation of UML class diagrams into graphs. The simple solution is transforming classes into nodes and generalizations and associations into edges. Still, if we want to convert, generalization sets, association classes, generalizations between associations, cardinalities, and several other constructs, this solution no longer works. Moreover, a challenge for the available frequent subgraph mining approaches is accounting for \textit{complex (semantically-rich)} graph data \cite{jiang2013survey,guvenoglu2018qualitative,song2022pattern}. 
There are several algorithms each of which admits graphs with different characteristics (e.g., labeled vs. unlabeled or directed vs. undirected). 


In the conceptual modeling context, we encounter construct-rich languages that challenge the capabilities of existing FSM solutions. For instance, some languages support multiple labels for classes and relations, multiple relationships between the same classes, and even classes of relations. This requires the mining algorithm to potentially handle various types of graphs, such as multi-labeled, directed, or multi-graphs. To the best of our knowledge, no existing algorithm fully accommodates all these graph types. Therefore, we ensured full compatibility with all algorithms by addressing the problem at its source. The importing process generates an LPG that captures all the information from conceptual models encoded in any conceptual modeling language, ensuring the output format is always suitable for the mining algorithm.

\begin{figure}[ht]
\centering
\renewcommand{\figurename}{Algorithm}%
\setcounter{figure}{0}%
\caption{Description of the generic importing algorithm.}
\colorbox[RGB]{249,249,249}{
\begin{minipage}{.80\linewidth}
\begin{algorithm}[H]
\footnotesize
\KwData{\texttt{Set of Conceptual Models $M$}}
\KwResult{\texttt{Set of Labeled Property Graphs (LPG) $G$}}
\For{each conceptual model $m \in M$}{
    \For{each concept $c$ in $m$}{
        map $c$ to a node $n_c$\;
        \If{$c$ represents a relation}{
            Connect $n_c$ to the source node with an edge $e_s$ labeled ``source''\;
            Connect $n_c$ to the target node with an edge $e_t$ labeled ``target''\;
        }
        \For{each property $p$ of $c$}{
            assign a label $l_p$ to node $n_c$\;
        }
    }
    create an undirected labeled graph $g$ using the set of nodes and edges\;
    add $g$ to $G$\;
}

\label{algorithm:0}
\end{algorithm}
\end{minipage}}
\end{figure}

Algorithm 1 presents a simplified overview of the importing process. The core idea is to \textit{reify} every element in the input conceptual model. For example, associations, cardinalities, and classes are all mapped into graph nodes with multiple labels. \autoref{importing} illustrates this transformation process, applied to three models encoded in different modeling languages: \textit{(a)} OntoUML, \textit{(b)} ArchiMate, and \textit{(c)} BPMN. On the right, the resulting LBGs are shown. The edge labels ``source'' and ``target'' are used to preserve the directionality of the relationships.

\begin{figure}[ht]
\centering
\setcounter{figure}{4}
\includegraphics[width=1\textwidth]{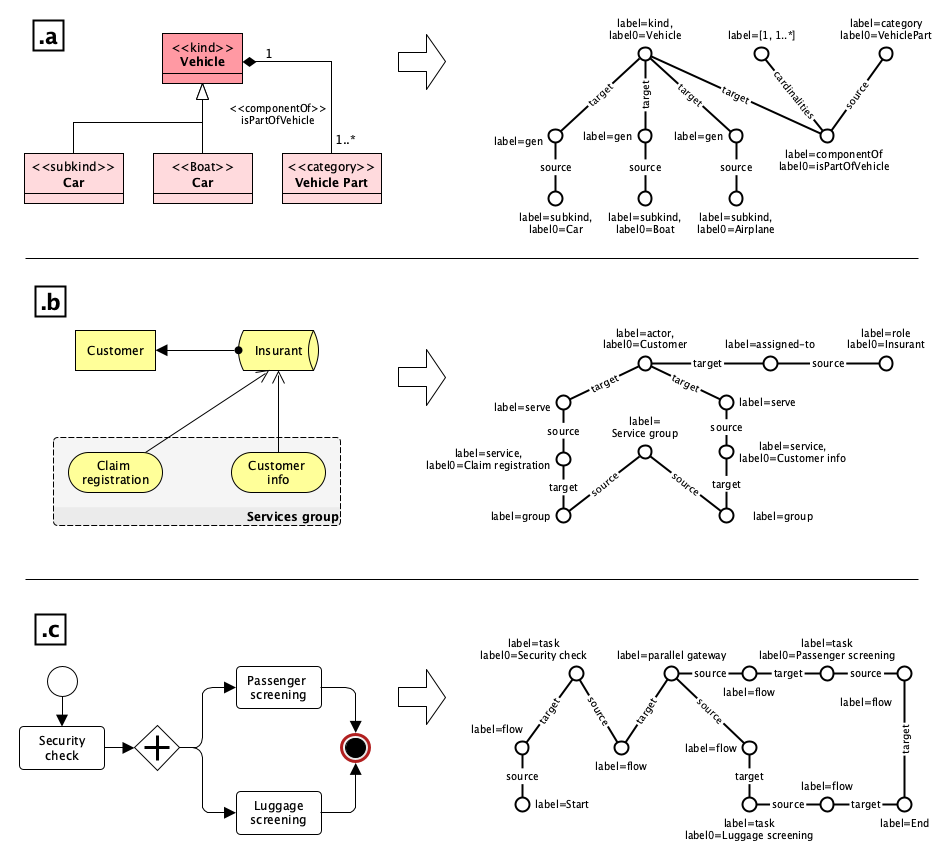}
\caption{Example of transformation from conceptual models to labeled property graphs.}
\label{importing}
\end{figure}

In the \textit{Filtering (1)} task, users can choose which language constructs to exclude from the models. For instance, with OntoUML, one may want to seek patterns only involving classes decorated with certain stereotypes, or involving only classes, generalizations, and generalization sets. Finally, the outputs of the filtering task are a set of (possibly filtered) LPGs (\textit{.ii} in \autoref{approach}) to be given as input to the mining algorithm and the storage of all the generated graphs (\textit{.iii}, which can be used to run the \textit{Deepening (3)} task (see \autoref{s:discovery} below).

\subsection{Discovering Structures Across and Within Models}
\label{s:discovery}

The ``Discovery'' phase is the workflow's core and aims to generate the candidate patterns in a format that can then be processed and made accessible to the user for the final assessment. This phase, in turn, is composed of two main tasks. 

\textit{Mining (2)}, the first task, involves applying an FSM algorithm. This allows for an interaction with the user, who can select:

\begin{enumerate}[label=\textit{.\roman*}]
\item the \textit{minimum structure size} for the output patterns (e.g., filter out patterns that have less than five nodes), 
\item the \textit{minimum frequency} threshold for the output patterns (e.g., filter out patterns that occur less than 30 times across models), and 
\item the \textit{known patterns} to be excluded from the final output (e.g., the user can provide graphs representing known patterns as input. These will be excluded from the final output to maximize the discovery of results that contain new information).
\end{enumerate}

The output of the mining task will then comprise a list of the discovered patterns in a format that eases the final assessment, along with a \textit{pattern index}, a list of indexes of the \textit{source models} in which the pattern occurs, and a \textit{model frequency value} calculated as the number of conceptual models that contain the pattern. For instance, given five models, the model frequency cannot be more than `5'.

An optional task named \textit{Deepening (3)} allows users to select a pattern from the previous task and discover the occurrences of this pattern within each input model. This operation can be run for each of the output patterns and implies using \textit{the selected pattern as a query to be run over the graphs storage generated via the importing task}. The output of the deepening task is a set of pattern occurrences for each selected pattern. This allows the user to derive the \textit{total frequency} for each pattern, calculated as the sum of the pattern occurrences within each input model (e.g., given five models, if the pattern occurs twice in each model the total frequency will be of `10', but the model frequency will be `5'). Through this step, the user can extract information that is related to the specific usage of the pattern within the reference model domain. For instance, via deepening, we can infer that, in a pattern, a node with stereotype \texttt{Kind} is associated with the label `Vehicle' in a model and `Means of transportation' in another model. 

\subsection{Assessing the Output Frequent Structures}

The ``Assessment'' phase aids users in analyzing the output. Similar to previous phases, this one comprises two main tasks. 

\begin{figure}[h!]
\centering
\renewcommand{\figurename}{Algorithm}%
\setcounter{figure}{1}%
\caption{Partial Description of the Clustering Algorithm.}
\colorbox[RGB]{249,249,249}{
\begin{minipage}{.70\linewidth}
\begin{algorithm}[H]
\footnotesize
\KwData{\texttt{Set of of Frequent Structures $F$}}
\KwResult{\texttt{Set of Clustered Frequent Structures $C$}}
\For{every frequent structure $f$, where $f\:\in\:F$}{
extract $\alpha$ as the number of nodes in $f$\;
extract $\beta$ the number of edges in $f$\;
extract the \textit{adjacency matrix} $A$ for $f$\;
\For{each node $n$ where $n\:\in\:f$}{
extract the associated labels $L$\;}
\For{$\alpha$, $\beta$, $A$ and $L$}
{flatten $A$ into a vector $v_A$ concatenating its rows\;
generate a vector $v_f=(\alpha,\beta,v_A,L)$\;
store $v_f$ in a list of vectors $V$\;}
}
\For{every vector $v_f^i\:\in V$}{
compute similarity with every other vector $v_f^j\:\in V$\;
take as input a similarity threshold $\gamma$\;
create the set of clusters $C$ according to $\gamma$\footnote{Note that, in our scenario, if we have three patterns, A, B, and C, and, according to a certain threshold, A is similar to B and B is similar to C, this implies that A is also similar to C, and A, B and C are in the same cluster.}\;
}
\label{algorithm:1}
\end{algorithm}
\end{minipage}}
\end{figure}

The \textit{Clustering (4)} task groups output structures---specifically, the recurrent or frequent structures from the mining task---based on their similarities.\footnote{Notice that in the approach we are proposing, to assess whether one conceptual model sub-structure is close to another, we take inspiration from conceptual model similarity techniques \cite{elkamel2016uml,ma2021two,guizzardi2022automated}, where the main task is to find similar models given a reference model input, to categorizing models according to their characteristics.} This should facilitate the user's process of scraping and consulting the output. Often, the number of outputs produced can surpass hundreds, and numerous structures can have similar characteristics. For example, some structures might differ by just one label (in one structure we might have three nodes where one is labeled as \texttt{Kind} and the other two specialize it as \texttt{Subkind}, while in another structure we might have three nodes where one is \texttt{Kind}, and the other two specialize it as \texttt{Subkind} and \texttt{Role}, respectively. In this sense, a cluster of patterns can be considered by the user as a set of variations, which sometimes may refer also to the same structural pattern (e.g., where a kind is specialized by two or three \texttt{Subkind}s). To enable clustering, in our process, we extract key characteristics from the graphs that represent each pattern. At present, the steps addressed to enable the clustering step are as from Algorithm 2. 

Currently, we adopt a single feature extraction method to calculate structure similarity. However, this does not prevent adopting different feature extraction or embedding techniques. This is valid also for the approach used to assess similarity, which now is limited to a \textit{cosine similarity} \cite{rahutomo2012semantic}, but in the future, it can be extended to other measures.

\begin{figure}[h!]
\centering
\setcounter{figure}{5}
\includegraphics[width=0.8\textwidth]{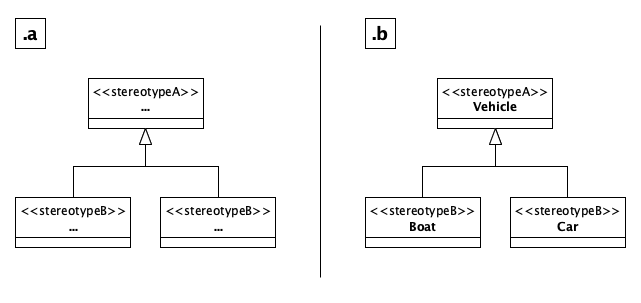}
\caption{Visualization example of recurrent structure \textit{(.a)} and a related occurrence \textit{(.b)}.}
\label{fig:vizexample}
\end{figure}

The \textit{Visualization (4)} task is crucial, focusing on creating an output that maximizes users' comprehension and engagement. This process revolves around transforming generated patterns, which inherently encapsulate the reified structure of input graphs, back into their original format --- a multi-directed graph. In this graphical representation, relationships, generalizations, and associations manifest as edges, complete with corresponding cardinalities to provide a richer context.

The significance of this visual representation lies in its ability to empower users in several ways. For example, by translating the outputs from the FSM algorithm into multi-directed graphs, users can better grasp complex relationships and hierarchies within the data. This enables deeper insights and understanding. \autoref{fig:vizexample} exemplifies the transformation power of visualization. On the left-hand side, you can observe a recurrent structure. On the right, you are presented with one of its occurrences, generated through the deepening task. This visual juxtaposition illustrates how the raw outputs of the mining and deepening tasks can be transformed into a visually accessible and actionable representation, empowering users to easily extract valuable knowledge from their data.

\section{Implementation}
\label{s:implementation}



We developed a command-line Python application for our approach, allowing users to interactively configure the process, select the input conceptual modeling language, manipulate the data, and assess the mining algorithm's output. The scripts we created are available at \href{https://github.com/unibz-core/CM-Mining}{\texttt{https://github.com/unibz-core/CM-Mining}}.

For tasks involving graph importing, processing, and transformation, we used the \textit{NetworkX library}\footnote{\href{https://networkx.org/}{\texttt{https://networkx.org/}}}, a powerful Python package for creating, analyzing, and visualizing complex networks. NetworkX played a crucial role in each step of the entire workflow. We integrated the \textit{Grandiso} library\footnote{\href{https://pypi.org/project/grandiso/}{\texttt{https://pypi.org/project/grandiso/}}}---a versatile Python library known for its efficient graph-matching and isomorphism-checking capabilities---in tasks involving graph matching, isomorphism detection, and motif search. For pattern mining, we relied on the \textit{gSpan Python} implementation\footnote{\href{https://pypi.org/project/gspan-mining/}{\texttt{https://pypi.org/project/gspan-mining/}}} of the frequent subgraph mining algorithm. This library enabled us to discover recurring subgraph patterns within our data. We achieved the visualization of discovered patterns using \textit{PlantUML}\footnote{\href{https://plantuml.com/}{\texttt{https://plantuml.com/}}}, a flexible tool for generating UML and ArchiMate diagrams. PlantUML is seamlessly integrated into our framework, enhancing the interpretability of our results through clear and informative visual representations.

This suite of tools enabled a robust and efficient framework for graph structure discovery and visualization
, ensuring we can efficiently access useful information from the mining outputs.

\subsection{OntoUML Miner}

As an initial proof of concept for our scientific contribution, we adapted our pipeline to support UML class diagrams, focusing specifically on OntoUML models. These represent the output of a trending paradigm situated at the confluence of conceptual modeling and ontology engineering, namely Ontology-driven conceptual modeling (ODCM). ODCM frequently entails the utilization of fundamental ontologies to steer the formulation of conceptual models, modeling languages, and tools \cite{verdonck2016insights}. Within this framework, the OntoUML modeling language~\cite{guizzardi2005ontological,guizzardi2021types}, has risen to prominence as one of the predominant methodologies \cite{verdonck2016insights}. 

\begin{figure}[ht]
\centering
\renewcommand{\figurename}{Algorithm}%
\setcounter{figure}{2}%
\caption{Partial Description of the OntoUML Importing Algorithm.}
\colorbox[RGB]{249,249,249}{
\begin{minipage}{.80\linewidth}
\begin{algorithm}[H]
\footnotesize
\KwData{\texttt{Set of of Conceptual Models $M$}}
\KwResult{\texttt{Set of Labeled Property Graphs (LPG) $G$}}
\For{every conceptual model $m$, where $m\:\in\:M$}{
map each class $c$ into a node $n_c$\;
map each association $a$ into a a node $n_a$\;
map each generalization $g$ into a node $n_g$\;
map association cardinalities $\phi$ into a node $n_\phi$\;
\For{each generalization node $n_g$}{
Connect $n_g$ to the parent class-node $n_c$ with an edge ``general''\;
Connect $n_g$ to the child class-node $n_c$ with an edge ``specific''\;}
\For{each association node $n_a$}{
Connect $n_a$ to the source class-node $n_c$ with an edge ``source''\;
Connect $n_a$ to the target class-node $n_c$ with an edge ``target''\;
\For{each cardinalities node $n_\phi$}{
Connect $n_\phi$ to the corresponding $n_a$ with an edge ``cardinalities''\;}
}
\For{each class $c$}{create the corresponding label(s) for node $n_c$\;}
\For{each association $a$}{create the corresponding label(s) for node $n_a$\;
}
create $LPG$ $g$\;
add a graph $g$ to $G$\;
}

\label{algorithm:2}
\end{algorithm}
\end{minipage}}
\end{figure}

\textit{OntoUML} extends the UML modeling language. Its meta-model is grounded on UFO (Unified Foundational Ontology) \cite{guizzardi2022ufo}, a formal theory based on contributions from Formal Ontology in Philosophy, Philosophical Logic, Cognitive Psychology, and Linguistics. UFO is one of the most used foundational ontologies in conceptual modeling and OntoUML is among the most used languages in ontology-driven conceptual modeling \cite{verdonck2016insights}. An example of core ontological distinctions underlying OntoUML is represented by the key categories of \textit{object types} (e.g., \texttt{Kind}, \texttt{Subkind}, \texttt{Role}s, and \texttt{RoleMixin}s), \textit{trope types} (e.g., \texttt{Relator}, \texttt{Mode}) and \texttt{Relation}s (\texttt{FormalRelations}, \texttt{MaterialRelations}, and \texttt{ParthoodRelations}).\footnote{For an in-depth analysis and characterization of the ontological categories underlying OntoUML, the reader is referred to \cite{guizzardi2005ontological,guizzardi2021types}.}

One of the main goals of OntoUML is to support the conceptual modeling activities by making explicit the semantics behind the modelers' design choices, thus enabling key features of the output conceptual models, such as understandability, interoperability, and reusability.

The purpose of the OntoUML miner is to enable the discovery of recurrent structures within OntoUML models, requiring the ability to handle all OntoUML constructs. To achieve this, the module adapts both the importing and visualization steps of the pipeline, which were identified as language-dependent in \autoref{approach}. The OntoUML-specific importing step is detailed in Algorithm 3, providing a specialized version of Algorithm 1. This partial view of the algorithm shows that the input concepts from the conceptual model correspond to specific OntoUML constructs, many of which also appear in UML. For example, taxonomic relations, associations, and relation properties like cardinalities are represented.\footnote{Note that the example pseudocode presented covers only a subset of constructs. Information on generalization sets, aggregation, and composition relations is not included. However, the importing step can readily handle these elements by applying the same reification strategy outlined in Algorithm 1 (e.g., a generalization set is represented as a node linked to generalization relation nodes, with properties such as disjoint and complete).}

As a final remark on the OntoUML visualization step, this process entails adapting the PlantUML transformation so that output patterns are displayed using OntoUML-like notation.

\subsection{ArchiMate Miner}

Besides OntoUML, we extended our pipeline to support ArchiMate models, demonstrating the adaptability of our mining approach to different modeling languages. ArchiMate, standardized by the Open Group,\footnote{\href{https://pubs.opengroup.org/architecture/archimate3-doc/}{\texttt{https://pubs.opengroup.org/architecture/archimate3-doc/}}} is one of the most widely used Enterprise Architecture (EA) modeling language~\cite{RoblB22}. The ArchiMate framework adopts a layered view of an enterprise, where the core entities of an enterprise are categorized along \textit{layers} (e.g., Business, Application, or Technology) and \textit{aspects} (e.g., Active Structure, Passive Structure, or Behavior). 

To enable the discovery of frequent structures within ArchiMate models, we adapted our pipeline to accommodate ArchiMate constructs during the language-dependent importing, filtering, and visualization steps.

The ArchiMate-specific importing step is detailed in Algorithm 4, extending the general approach outlined in Algorithm 1. This step transforms ArchiMate models into Labeled Property Graphs (LPGs) by mapping elements and relationships to nodes and assigning labels that capture their name and type information (e.g., \texttt{BusinessProcess} for an element or \texttt{Assignment} for a relationship). Specialization relationships are treated distinctly, with nodes connected by edges labeled \texttt{general} and \texttt{specific} to ensure that hierarchical structures are distinctly represented and allow for additional filtering.

The filtering step allows users to refine the models before mining, reducing the graph size and, subsequently, the search space. Users can apply filters based on:

\begin{enumerate}[label=\textit{.\roman*}]
\item element types (e.g., \texttt{ApplicationComponent}, \texttt{TechnologyService}), 
\item layers or aspects (e.g., focus solely on the Technology Layer or Behavior Aspect),
\item relationship types (e.g., include only \texttt{Realization} or \texttt{Access} relationships), and 
\item edge labels (e.g., filtering by \texttt{general}/\texttt{specific} or \texttt{source}/\texttt{target} labels).
\end{enumerate}
\vspace{0.5em}

Finally, similar to the OntoUML implementation, the visualization step was adapted to render ArchiMate patterns using PlantUML, which natively supports ArchiMate diagrams.\footnote{\href{https://plantuml.com/archimate-diagram}{\texttt{https://plantuml.com/archimate-diagram}}}

\begin{figure}[ht]
\centering
\renewcommand{\figurename}{Algorithm}%
\setcounter{figure}{3}%
\caption{Partial Description of the ArchiMate-specific Importing Algorithm.}
\colorbox[RGB]{249,249,249}{
\begin{minipage}{.90\linewidth}
\begin{algorithm}[H]
\footnotesize
\KwData{\texttt{Set of of ArchiMate Models $M$}}
\KwResult{\texttt{Set of Labeled Property Graphs (LPG) $G$}}
\For{every model $m$, where $m\:\in\:M$}{
map each element $e$ into a node $n_e$\;
map each relationship $r$ into a a node $n_r$\;
map each specialization $s$ into a node $n_s$\;
\For{each specialization node $n_s$}{
Connect $n_s$ to the parent element-node $n_e$ with an edge ``general''\;
Connect $n_s$ to the child element-node $n_e$ with an edge ``specific''\;}
\For{each relationship node $n_r$}{
Connect $n_r$ to the source element-node $n_e$ with an edge ``source''\;
Connect $n_r$ to the target element-node $n_e$ with an edge ``target''\;
}
\For{each element $e$}{create the corresponding label(s) for node $n_e$\;}
\For{each relationship $a$}{create the corresponding label(s) for node $n_r$\;
}
create $LPG$ $g$\;
add graph $g$ to $G$\;
}
\end{algorithm}
\label{algorithm:archi-import}
\end{minipage}}
\end{figure}

\section{Experiments}
\label{s:eval}
In this section, we test our approach through three experiments, keeping as reference the requirements described in \autoref{s:requirements}, and addressing the following research questions:

\vspace{0.5em}
\begin{itemize}
 \setlength{\itemindent}{1.0em}
    \item [\textbf{RQ1}] 
    \textit{Can the proposed approach generate structures encoding previously recognized interesting patterns?}
    This research question is aimed at testing whether the proposed solution can discover pre-identified interesting patterns \textbf{(R1)}. This research question is also used to check the role of the customization steps in supporting the discovery process \textbf{(R2)} and the level of comprehensibility of the outcome \textbf{(R3)}.
    \item [\textbf{RQ2}] \textit{What are the main parameters affecting the performance of the approach?} Here, we want to identify the characteristics of the input data or parameters having a major influence on the performance \textbf{(R4)}.
    \item [\textbf{RQ3}] \textit{Is the clustering step accurate in grouping structures?} This research question is mainly concerned with \textbf{(R3)}. Here, we want to assess the practical utility of a key component in the presentation of the output. 
\end{itemize}

\vspace{0.5em}

The data about the experiments and detailed instructions for reproducibility are available at this \texttt{GitHub} link: \href{https://github.com/unibz-core/cmining-approach}{\texttt{https://github.com/unibz-core/cmining-approach}}.

\subsection{Experiment 1: Reliability Test}

The primary objective of this experiment is to validate the approach with respect to \textbf{RQ1}. Specifically, we aim to determine whether the choices made in the importing task enable the discovery of structures similar to those previously identified by domain experts. To achieve this, we used known patterns for each language as a reference and observed the impact of selected input parameters on the mining task outputs. For full control over the validation, we created ten models, each containing instances of the known patterns. This controlled context and small dataset also allowed us to closely monitor how filtering actions effectively prune results that fall outside the search scope.

\subsubsection{Using the OntoUML Dataset}


This experiment used an \textit{ad hoc} set of models with some OntoUML patterns we already know as input and we checked how many expected patterns were found.

\vspace{0.5em}

\noindent\textbf{Data:} A dataset comprising 10 models developed by the authors of this paper. These are small models (varying from a minimum of \textit{6 classes} and \textit{5 relations} to a maximum of \textit{10 classes} and \textit{8 relations})\footnote{The model files are available here: \href{https://github.com/unibz-core/cmining-approach/tree/main/ontouml/experiment1}{\texttt{cmining-approach/tree/main/ontouml/experiment1}} } that were created specifically for this experiment, taking inspiration from structures that are present in real models, with the goal of reproducing a controlled number of target patterns. The distribution of patterns per model can be observed in \autoref{patternsDistribution}.

\vspace{0.5em}
\noindent\textbf{Setup:} For validation, we used six common OntoUML patterns that served as ``litmus test''.\footnote{A litmus test is ``a critical indicator of future success or failure'' A is a litmus test for B if A can be effectively used to measure some property of B \cite{litmus}.} These patterns were previously manually identified as useful for building OntoUML models by the designers of the language within multiple example models \cite{guizzardi2014ontological,ruy2017reference}. We represent the selected patterns in \autoref{ontoUMLpatterns}.

\begin{figure}[t]
\centering
\setcounter{figure}{6}
\includegraphics[width=1\textwidth]{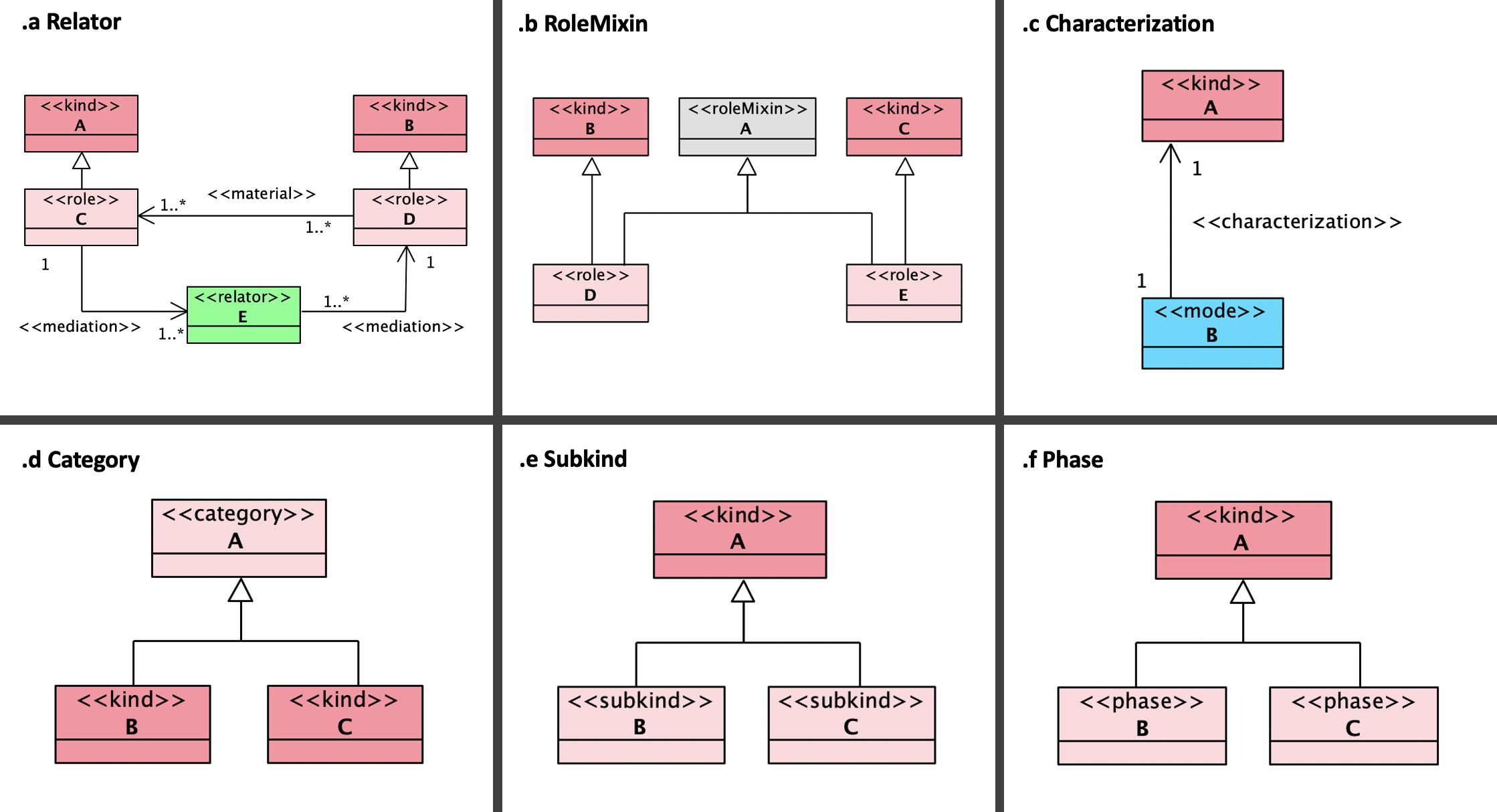}
\caption{OntoUML modeling patterns examples \cite{ruy2017reference}.}
\label{ontoUMLpatterns}
\end{figure}

We executed the application seven times (one without adopting parameters to filter specific concepts and the other times with six different configurations, each to find one of the 6 selected target patterns) and we checked whether the proposed solution could discover the pre-identified interesting patterns (\textbf{R1}). Moreover, we tested the role of the customization steps in supporting the discovery process (\textbf{R2}) and the level of comprehensibility of the outcome (\textbf{R3}). First, we conducted a trial adopting no customization facility from the pipeline. As parameters, we selected 3 (the support of the less frequent patterns, i.e., \texttt{Relator}, \texttt{Subkind}, and \texttt{Phase}) as minimum support, and 5 as minimum number of nodes (the size of the smaller pattern, i.e., \texttt{Characterization}),\footnote{Where we have 2 nodes for the classes, 1 node as an association and 2 nodes for the cardinalities.} so that all patterns can be discovered. For each other trial, we customized the pipeline to find the target pattern, e.g., by filtering some constructs and some other patterns we did not want to discover. For instance, for the \texttt{Relator} pattern, we selected \texttt{Kind}, \texttt{Relator}, and \texttt{Role} as target constructs and we filtered out information about the generalizations sets (the customization we adopted in terms of \textit{minimum support}, \textit{nodes number}, and \textit{selected/removed constructs} is reported in \autoref{ex1output}.).\footnote{To provide context on the OntoUML constructs present in the patterns, here is a brief explanation: A \texttt{Kind} is a construct commonly used across models to represent rigid concepts that establish an identity principle. A \texttt{Subkind} represents rigid specializations of identity providers, such as \texttt{Kind}s, while the \texttt{Relator} construct is used to represent \textit{truth-makers} of \textit{material relations—entities} that must exist for two or more individuals to be connected by \textit{material relations}. A \texttt{Role} represents anti-rigid specializations of identity providers, like \texttt{Kind}s. The \texttt{Category} construct serves as a \textit{rigid mixin}, which does not depend on a specific identity principle but is used to aggregate essential properties across individuals with different identity principles. A \texttt{RoleMixin} is the equivalent of \texttt{Role} for types that aggregate instances with different identity principles. The \texttt{Phase} stereotype represents anti-rigid subtypes of identity providers, such as \texttt{Kind}s, that arise due to changes in intrinsic properties (for example, a person’s age). Finally, \texttt{Characterization} is a relation that connects a bearer type with its features. More detailed information on OntoUML constructs can be found in \cite{guizzardi2005ontological} and \href{https://github.com/OntoUML/ontouml-models}{\texttt{https://github.com/OntoUML/ontouml-models}}.}

\begin{table}[h]
\centering
\caption{Patterns distribution over the synthetic data set.}
\label{patternsDistribution}
\begin{tabularx}{\textwidth}{lXXXXXX}
\textbf{Model}         & \textbf{\texttt{Relator}} & \textbf{\texttt{RoleMixin}} & \textbf{\texttt{Charact.}} & \textbf{\texttt{Category}} & \textbf{\texttt{Subkind}} & \textbf{\texttt{Phase}} \\ 
\toprule
01                     & 0                & 0                  & 1                         & 1                 & 1                & 1              \\
\midrule
02                     & 0                & 0                  & 2                         & 1                 & 0                & 1              \\
\midrule
03                     & 0                & 0                  & 1                         & 1                 & 0                & 2              \\
\midrule
04                     & 1                & 0                  & 1                         & 0                 & 0                & 0              \\
\midrule
05                     & 1                & 0                  & 1                         & 1                 & 0                & 0              \\
\midrule
06                     & 0                & 1                  & 0                         & 2                 & 0                & 0              \\
\midrule
07                     & 0                & 1                  & 1                         & 0                 & 1                & 0              \\
\midrule
08                     & 1                & 0                  & 1                         & 1                 & 0                & 0              \\
\midrule
09                     & 0                & 1                  & 0                         & 2                 & 0                & 0              \\
\midrule
10                     & 0                & 1                  & 1                         & 0                 & 1                & 0  
\\ 
\midrule
\midrule
\textit{Overall Freq.} & 3                & 4                  & 9                         & 9                 & 3                & 4              \\
\midrule
\textit{Model Freq.}   & 3                & 4                  & 8                         & 7                 & 3                & 3 \\ 
\bottomrule
\end{tabularx}
\end{table}


\begin{table}
\renewcommand{\arraystretch}{1.2}
\centering
\caption{Discovered patterns. ``Neutral'' is the trial where no customization has occurred. Each of the other records represents a trial performed to find a target pattern. For instance, ``Relator'' was performed to find the corresponding pattern and the total patterns found were 42.}
\label{ex1output}
\begin{minipage}{\linewidth}
\begin{tabularx}{\textwidth}{lcclc}
\textbf{trial}            & \multicolumn{1}{l}{\textbf{freq.}} & \multicolumn{1}{l}{\textbf{nodes}} & \textbf{filter}                                                                 & \multicolumn{1}{l}{\textbf{freq. strs.}} \\ 
\toprule
\texttt{Neutral}          & 3       & 5     & \textit{none}                                                                   & 4045     \\ 
\midrule
\texttt{Relator}          & 3       & 12    & \{  \texttt{Select: Kind, Role, Relator} \}  \{ \texttt{Remove: Charact}  \}          & 42       \\ 
\midrule
\texttt{RoleMixin}        & 4       & 10    & \{ \texttt{Select: Kind, Role, Rmixin} \} \{ \texttt{Remove: Assoc} \} & 2        \\ 
\midrule
\texttt{Charact.} & 8       & 4     & \{ \texttt{Select: Kind, Mode} \} \{\texttt{Remove: Gen} \}                     & 1\footnote{Note that the result ``1'' does not mean that only one occurrence of that pattern was found, but that exactly one structure corresponding to that pattern was found. This structure, in turn, can occur multiple times, for example, in this case, the frequency was ``8'', as expected.}        \\
\midrule
\texttt{Category}         & 6       & 4     & \{  \texttt{Select: Kind, Category} \} \{ \texttt{Remove: Assoc} \}         & 32       \\ 
\midrule
\texttt{Subkind}          & 3       & 4     & \{ \texttt{Select: Kind, Subkind} \} \{ \texttt{Remove: Assoc} \}          & 41       \\ 
\midrule
\texttt{Phase}            & 3       & 4     & \{ \texttt{Select: Kind, Phase} \} \{ \texttt{Remove: Assoc} \}             & 89       \\
\bottomrule
\end{tabularx}
\end{minipage}
\end{table}

\vspace{0.5em}
\noindent\textbf{Results:} The conducted trial successfully identified all the target patterns, demonstrating the pipeline's proficiency in addressing \textbf{R1}. The approach identified all the (classes of) patterns that were explicitly identified \textit{a priori} and correctly counted their occurrences.

The ``Neutral'' trial yielded an extensive list of output patterns, numbering in the thousands, augmenting the challenge of pinpointing the target pattern.

In subsequent trials, the application of the customization features significantly decreased the number of output patterns, thus showing the key role of the features we implemented in the discovery process (\textbf{R2}). For instance, with the \texttt{Relator} pattern, we selectively filtered extraneous stereotypes classes and relations out, resulting in a noteworthy reduction of the number of outputs (from 4045 to 42, see \autoref{ex1output}). We provide a visual representation and explain the \texttt{Relator} pattern in \autoref{fig:ex1r}. Particularly, the pipeline accurately and comprehensively identified all constructs originating from the input graphs.

\begin{figure}[h!]
\centering
\includegraphics[width=1\textwidth]{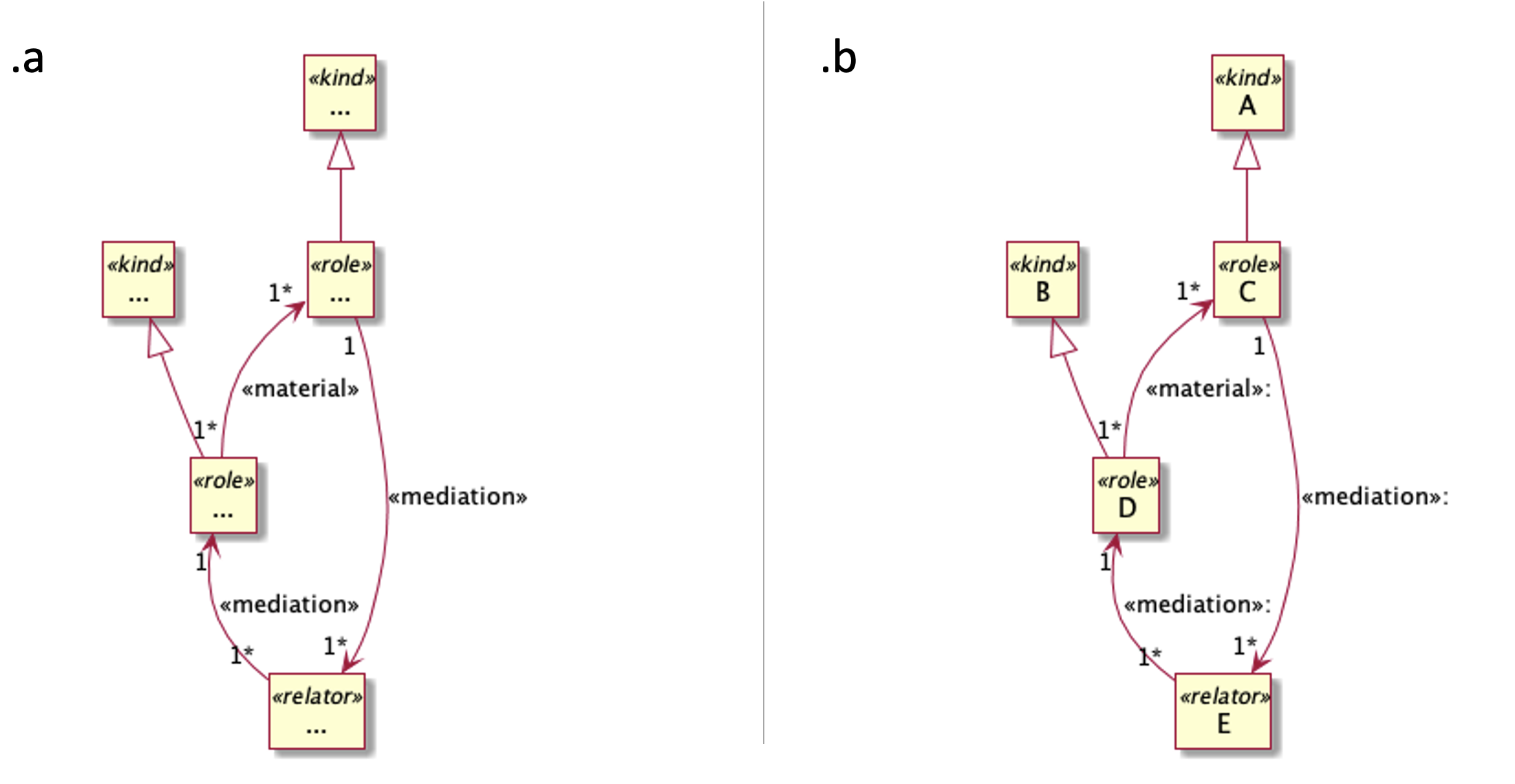}
\caption{Example of visualization for a \texttt{Relator} pattern occurrence as the pipeline returned it.}
\label{fig:ex1r}
\end{figure}


\subsubsection{Using the ArchiMate Dataset}
\label{sec:ex1-archimate}

Similarly to the OntoUML experiment, we used an \textit{ad hoc} set of ArchiMate models, containing various known patterns as input, and we conducted different trials to verify that the expected patterns can be found in the mining output.

\noindent\textbf{Data:} This experiment used a synthetic dataset of 10 manually created ArchiMate models, with sizes ranging from 10 to 15 elements and 8 to 12 relationships. Six EA Smells were selected as target anti-patterns for this analysis, as illustrated in \autoref{archimatePatterns}. These patterns include \texttt{Chatty Service} (\texttt{CS}), \texttt{Combinatorial Explosion} (\texttt{CE}), \texttt{Cyclic Dependency} (\texttt{CD}), \texttt{Data Service} (\texttt{DS}), \texttt{Multifaceted Abstraction} (\texttt{MA}), and \texttt{Wrong Cuts} (\texttt{WC}). EA Smells~\cite{SalentinH20} serve as qualitative indicators of structural inefficiencies and represent potential issues that affect the non-functional aspects of EA models (e.g., maintenance). Analogous to code smells that signal technical debt in source code, EA Smells assesses an organization holistically, beyond purely technical scopes~\cite{SalentinH20}. The EA Smells used in this experiment were selected from an extensive catalog\footnote{\href{https://swc-public.pages.rwth-aachen.de/smells/ea-smells/}{\texttt{https://swc-public.pages.rwth-aachen.de/smells/ea-smells/}}} and translated into ArchiMate patterns (cf.~\cite{SmajevicHB21}) using the provided descriptions. The distribution of the patterns across the dataset is listed in \autoref{patternsDistribution-archimate}.

\begin{figure}[!ht]
\centering
\setcounter{figure}{8}
\includegraphics[width=1\textwidth]{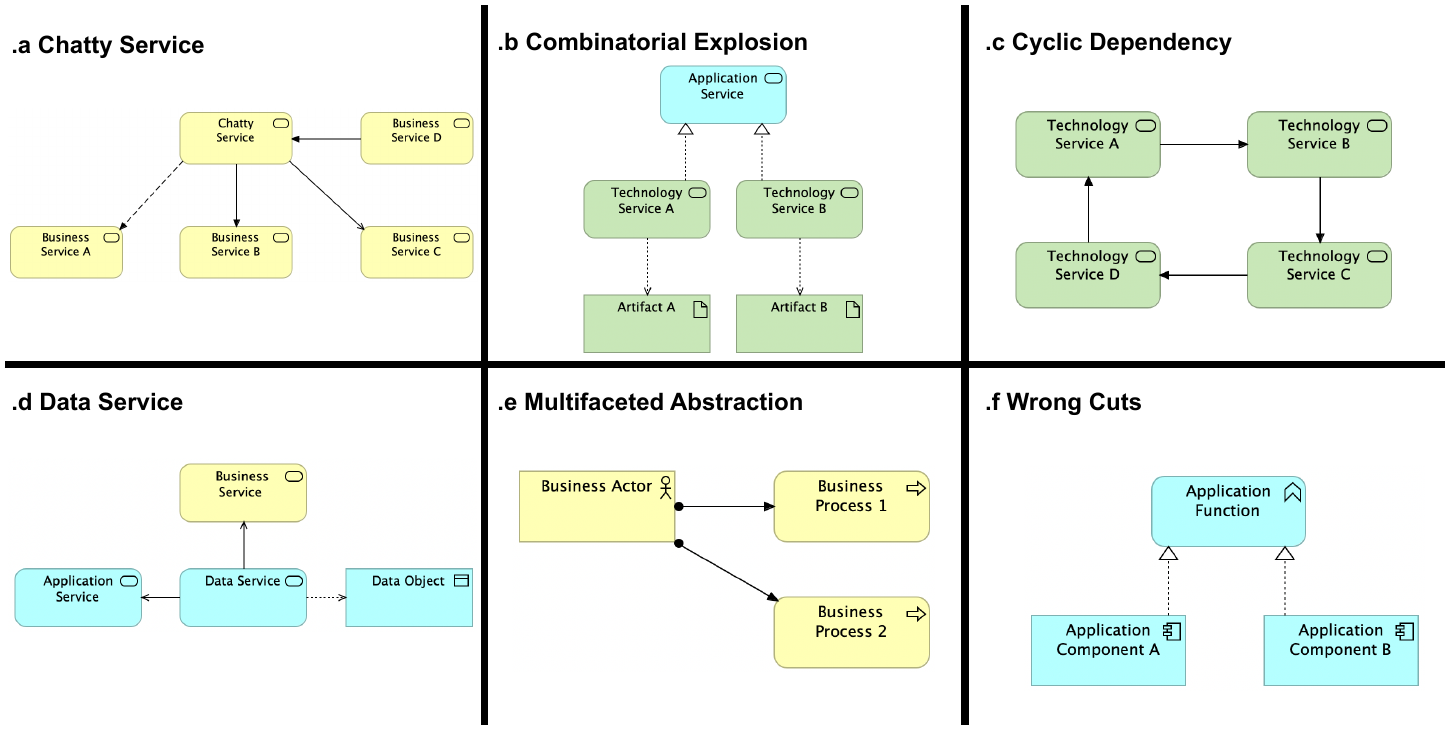}
\caption{ArchiMate selected EA Smells Patterns.}
\label{archimatePatterns}
\end{figure}

\begin{table}[!ht]
    \centering
    \caption{Patterns distribution over the synthetic ArchiMate dataset}
    \label{patternsDistribution-archimate}
    \begin{tabularx}{\textwidth}{lXXXXXX}
        \textbf{Model} & \texttt{CS} & \texttt{CE} & \texttt{CD} & \texttt{DS} & \texttt{MA} & \texttt{WC} \\ 
        \toprule
        01 & 1 & 0 & 1 & 0 & 1 & 0 \\ 
        \midrule
        02 & 0 & 1 & 1 & 0 & 0 & 2 \\ 
        \midrule
        03 & 0 & 1 & 0 & 1 & 0 & 1 \\ 
        \midrule
        04 & 1 & 0 & 0 & 1 & 1 & 0 \\ 
        \midrule
        05 & 0 & 1 & 0 & 0 & 1 & 2 \\ 
        \midrule
        06 & 0 & 0 & 2 & 0 & 0 & 1 \\ 
        \midrule
        07 & 0 & 0 & 1 & 0 & 1 & 1 \\ 
        \midrule
        08 & 1 & 0 & 1 & 1 & 0 & 0 \\ 
        \midrule
        09 & 0 & 1 & 0 & 0 & 2 & 1 \\ 
        \midrule
        10 & 0 & 0 & 2 & 0 & 0 & 1 \\ 
        \midrule
        \midrule
        \textit{Overall Freq.} & 3 & 4 & 8 & 3 & 6 & 9 \\
        \textit{Model Freq.} & 3 & 4 & 6 & 3 & 5 & 7 \\
        \bottomrule
    \end{tabularx}
\end{table}

\vspace{0.5em}
\noindent\textbf{Setup:} The experiment was organized into seven trials. The first, referred to as the  ``neutral'' trial, mined patterns without applying any filters to capture all potential patterns. Parameters were set to identify patterns with a frequency of at least 3 (matching the frequency of less frequent patterns, e.g., \texttt{Chatty Service} and \texttt{Data Service}) and a minimum node count of 5 (corresponding to the size of the smallest patterns, e.g., \texttt{Multifaceted Abstraction} and \texttt{Wrong Cuts}).

The subsequent six trials focused on identifying the individual patterns by applying specific filters to narrow the search space. For instance, \texttt{Cyclic Dependency} patterns in the dataset exclusively consist of \texttt{TechnologyService} elements and \texttt{Triggering} relationships. Hence, filters were applied to include only these elements and relationships while excluding others. The mining parameters, such as the minimum frequency and node count, were also adjusted to ensure patterns of interest were captured. The complete configuration of parameters and filters for each trial is summarized in \autoref{ex1output-archimate}.

\begin{figure}[!ht]
\centering
\includegraphics[width=0.9\textwidth]{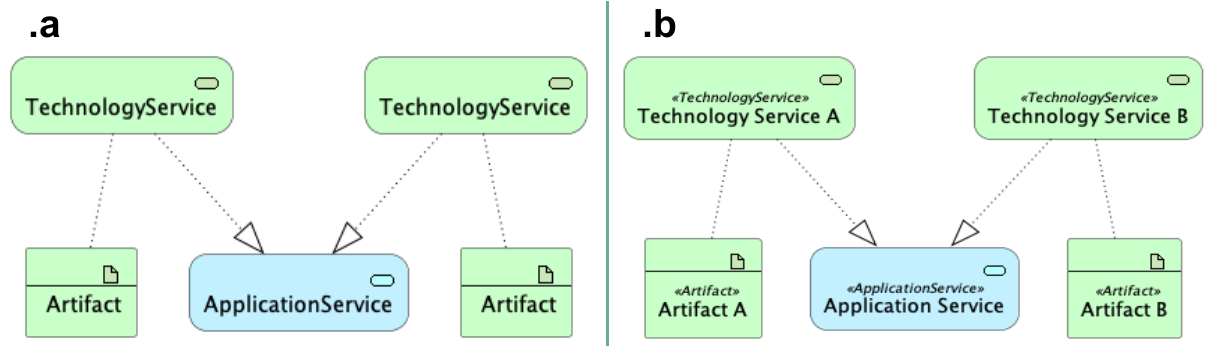}
\caption{Example visualization for a \texttt{Combinatorial Explosion} pattern and occurrence as the pipeline returned it: \textbf{.a} shows the pattern with its related element- and relationship types, while \textbf{.b} shows an example occurrence of the pattern including element names.}
\label{fig:ex1-archi-viz}
\end{figure}

\vspace{0.5em}
\noindent\textbf{Results:} Table~\ref{ex1output-archimate} summarizes the results of the seven trials. The neutral trial yielded 80 total patterns, including unrelated or redundant outputs, showcasing the need for customized filtering to refine results. In contrast, the filtered trials successfully identified all target patterns, with each trial isolating the intended structure and significantly reducing the number of output patterns (\textbf{R2}). For instance, the \texttt{Wrong Cuts} (\texttt{WC}) pattern, distributed 9 times across 7 models (see \autoref{patternsDistribution-archimate}), was correctly identified as a single recurring pattern during the corresponding trial, when applying the specified parameters and filters. Similarly, all other filtered trials correctly identified the intended EA Smell patterns, confirming the approach's reliability in detecting known structures and accurately counting their occurrences (\textbf{R1}). Note that in the \texttt{Cyclic Dependency} (\texttt{CD}) trial 3 patterns were found, consisting of our target pattern and two duplicates, due to the cycle in the graph structure. An example visualization for a concrete pattern, as returned from the pipeline, is shown in \autoref{fig:ex1-archi-viz}.

\begin{table}[!ht]
    \centering
    \caption{Discovered Patterns}
    \label{ex1output-archimate}
    \begin{tabularx}{\textwidth}{lccXc}
        \toprule
        \textbf{trial} & \textbf{freq} & \textbf{nodes} & \textbf{filter} & \textbf{freq. strs.} \\ 
        \midrule
        \texttt{Neutral} & 3 & 5 & \textit{none} & 80 \\ 
        \midrule
        \texttt{CS} & 3 & 9 & \{ \texttt{BusinessProcess} \} \{ \texttt{Flow}, \texttt{Serving}, \texttt{Triggering} \} & 1 \\ 
        \midrule
        \texttt{CE} & 4 & 9 & \makecell[l]{ \{ \texttt{TechnologyService}, \texttt{Artifact}, \texttt{ApplicationService} \} \\
        \{ \texttt{Realization}, \texttt{Access} \} }& 1 \\ 
        \midrule
        \texttt{CD} & 6 & 8 & \{ \texttt{TechnologyService} \} \{  \texttt{Triggering} \} & 3 \\ 
        \midrule
        \texttt{DS} & 3 & 7 & \makecell[l]{ \{ \texttt{ApplicationService}, \texttt{BusinessService}, \texttt{DataObject} \} \\
        \{ \texttt{Access}, \texttt{Serving} \} } & 1 \\ 
        \midrule
        \texttt{MA} & 5 & 5 & \{ \texttt{BusinessActor}, \texttt{BusinessProcess} \} \{ \texttt{Assignment} \} & 1 \\ 
        \midrule
        \texttt{WC} & 7 & 5 & \makecell[l]{ \{ \texttt{ApplicationComponent}, \texttt{ApplicationFunction} \} \\
        \{ \texttt{Realization} \} } & 1 \\ 
        \bottomrule
    \end{tabularx}
\end{table}

\subsubsection{Threats to Validity}

This experiment focused on a limited set of patterns that, although relatively intricate, represent only a fraction of a broader spectrum of patterns. Moreover, the synthetic datasets used in this study were fairly compact. However, the combination of the patterns we selected from the OntoUML and the ArchiMate scenarios is suitable for assessing whether our approach can discover structures with the required level of expressiveness.

\subsection{Experiment 2: Performance Test}

This experiment primarily addresses \textbf{RQ2}. Here, performance is not measured by the number of patterns identified versus those expected, as the chosen mining algorithm has been verified to be 100\% accurate \cite{yan2002gspan}, making metrics like precision and recall irrelevant. Likewise, the main focus is not the time required to generate output, given that FSM tasks are known to be time-consuming. Instead, the goals are: \textit{.i} to assess whether the algorithm can produce results on real datasets containing models of varying sizes when run on basic, affordable hardware, and \textit{.ii} to observe how the execution time scales with an increase in input models.

It is important to note that performance is influenced by the specific implementation choices made for each module in the pipeline, which can vary. For instance, future versions of the application may adopt a different mining algorithm if needed. This highlights the modular nature of our approach, which allows different components to be interchanged to support the same tasks effectively.

\subsubsection{Using the OntoUML Dataset}
\label{UMLperformance}

\noindent\textbf{Data:} The input data comprised 94 models from a catalog of \textit{OntoUML} models \cite{sales2023fair}. The catalogue results from a community effort, which collected models of different sizes, describing different domains, and developed for varying purposes in different contexts.

\vspace{0.5em}
\noindent\textbf{Setup:} This validation was executed on a \textit{MacBook Pro} (Retina, 13-inch, Early 2015) with CPU 2,7 GHz Intel Core i5 and was organized in two main trials: a baseline \textit{trial (0)}, where we selected as input the 47 conceptual models from the OntoUML dataset and \textit{trial (1)} where we added 47 more models. The selected models have different sizes, in terms of relations and and classes. As an example, the graphs generated from these sources can have a small number of nodes and relations, namely 47 and 35 respectively (see \texttt{lindeberg2022simple-ontorights.json}, in the reference experiments repository), or higher, namely 2.449 and 2.282 (see \texttt{indeberg2022full-ontorights.json}). For each step, to increase the amount of data to be handled, we also tested the approach with six different customizations (we show these values in \autoref{perf}) concerning partition size parameters and the minimum frequency of the mining task.

\vspace{0.5em}
\noindent\textbf{Results:} \autoref{perf} summarizes the results of this experiment by providing the time taken by the graph partitioning, the mining step, the clustering, and the generation of a visual representation for each pattern. The execution time reveals that the processing time \textit{rises} when:

\begin{enumerate}[label=\textit{.\roman*}]
\item the number of models increases, 
\item the number of nodes for the partitioned graph decreases, and 
\item the minimum frequency threshold is decreased, thus allowing to find a larger number of patterns. 
\end{enumerate}

As we might expect, the function that takes the most time is the one dedicated to mining and pattern generation. 

\begin{table}[h!]
\centering
\caption{Experiment 2 results. For each of the two trials (47 and 94 models), we selected the same parameters (nodes and frequency) and provided the number of patterns plus the time taken by four functions of the pipeline.}
\begin{tabular}{cccccccc}
\toprule
\textbf{models} & \textbf{import (s)} & \textbf{nodes} & \textbf{freq.} & \textbf{mining (s)} & \textbf{patterns} & \textbf{clustering (s)} & \textbf{viz. (s)} \\
\midrule
\multirow{6}{*}{47}                 & \multirow{6}{*}{0.7327}                    & 5                                  & 20                                     & 1.375      & 1                                     & 0.011                              & 0.106                        \\
                                    &                                            & 3                                  & 20                                     & 1.591      & 15                                    & 0.043                              & 1.648                        \\
                                    &                                            & 5                                  & 15                                     & 3.792      & 22                                    & 0.073                              & 2.257                        \\
                                    &                                            & 3                                  & 15                                     & 3.950      & 65                                    & 0.422                              & 6.669                        \\
                                    &                                            & 5                                  & 10                                     & 47.579     & 246                                   & 0.982                              & 16.387                       \\
                                    &                                            & 3                                  & 10                                     & 48.448     & 372                                   & 1.360                              & 20.387                       \\ 
                                    \midrule
\multirow{6}{*}{94}                 & \multirow{6}{*}{1.238}            & 5                                  & 20                                     & 130.458    & 37                                    & 0.103                              & 3.781                        \\
                                    &                                            & 3                                  & 20                                     & 184.239    & 72                                    & 0.503                              & 7.360                        \\
                                    &                                            & 5                                  & 15                                     & 224.782    & 141                                   & 0.734                              & 9.410                        \\
                                    &                                            & 3                                  & 15                                     & 229.138    & 193                                   & 0.812                              & 11.500                       \\
                                    &                                            & 5                                  & 10                                     & 324.781    & 586                                   & 1.990                              & 25.675                       \\
                                    &                                            & 3                                  & 10                                     & 359.318    & 662                                   & 3.156                              & 31.057                       \\ \bottomrule
\end{tabular}

\label{perf}
\end{table}


\subsubsection{Using the ArchiMate Dataset}

\noindent\textbf{Data:} For this experiment, we used subsets of 50 and 100 medium-sized ArchiMate models, extracted from the EAModelSet dataset~\cite{GlaserSB23}. The EAModelSet is a FAIR dataset comprising over 900 ArchiMate models of varying sizes, collected from GitHub, GenMyModel, and the EA community. From the dataset, we first filtered for English-language models and then selected medium-sized models with 30 to 80 relationships each. The 50-model subset contained 2.862 elements and 3.570 relationships, while the 100-model subset comprised 5.890 elements and 6.920 relationships.

\vspace{0.5em}
\noindent\textbf{Setup:} 
The experiment was run on the machine adopted also in the experiment with OntoUML models (see \autoref{UMLperformance})  
Two primary trials were conducted: one with 50 models and the other with 100 models. For each trial, six different parameter combinations were tested, progressively increasing the amount of data to be processed. No filters were applied, meaning all elements and relationships in the models were analyzed. A timeout of 600 seconds (10 minutes) was imposed for the mining step to ensure execution feasibility. The parameter combinations and results are presented in \autoref{ex2output-archimate}.

\vspace{0.5em}
\noindent\textbf{Results:} \autoref{ex2output-archimate} summarizes the results, detailing the time taken for each step of the pipeline: graph partitioning (import), mining, clustering, and visualization. For the visualization step, only the time required to generate PlantUML.txt files was measured. As expected, the mining step accounted for most of the execution time. The results show that processing time increased under the following conditions:

\begin{enumerate}[label=\textit{.\roman*}]
\item when the number of total elements and relationships is increased (e.g., moving from 50 to 100 models),
\item when the minimum node parameter decreased, allowing smaller patterns to be considered,
\item when the frequency threshold was reduced, resulting in a larger number of patterns being mined. 
\end{enumerate}

For the 50-model subset, all parameter combinations were completed successfully within the timeout limit. However, for the 100-model subset, the mining step exceeded the 600-second timeout in both trials with the lowest frequency (10), indicating that processing models of this size with such settings would require more time and more robust hardware. These results reaffirm the high computational cost of the mining step, particularly when handling larger datasets with more lenient parameters. The results also show the need to balance parameter settings with hardware capabilities to ensure efficient execution.

\begin{table}[h!]
\centering
\caption{Experiment 2 results with the ArchiMate dataset}
\begin{tabular}{cccccccc}
\toprule
\textbf{models} & \textbf{import (s)} & \textbf{nodes} & \textbf{freq.} & \textbf{mining (s)} & \textbf{patterns} & \textbf{clustering (s)} & \textbf{viz. (s)} \\
\midrule
\multirow{6}{*}{50} & \multirow{6}{*}{0.066} 
& 5 & 20 & 0.178 & 1 & 0.545 & 0.153 \\
 & & 3 & 20 & 0.182 & 11 & 0.562 & 0.137 \\
 & & 5 & 15 & 1.257 & 42 & 0.571 & 0.124 \\
 & & 3 & 15 & 1.283 & 79 & 0.582 & 0.147 \\
 & & 5 & 10 & 66.810 & 502 & 0.593 & 0.159 \\
 & & 3 & 10 & 87.120 & 604 & 0.624 & 0.187 \\
\midrule
\multirow{6}{*}{100} & \multirow{6}{*}{0.154} 
& 5 & 20 & 125.447 & 8 & 0.657 & 0.201 \\
 & & 3 & 20 & 131.459 & 38 & 0.683 & 0.207 \\
 & & 5 & 15 & 455.807 & 68 & 1.032 & 0.314 \\
 & & 3 & 15 & 507.853 & 93 & 1.154 & 0.462 \\
 & & 5 & 10 & 600+ & - & - & - \\
 & & 3 & 10 & 600+ & - & - & - \\
\bottomrule
\end{tabular}
\label{ex2output-archimate}
\end{table}

\subsubsection{Threats to Validity}

The primary challenge to the validity of this experiment stems from the absence of a comparative analysis involving multiple devices. The ultimate performance might be influenced by additional variables not present in the current setting. Nonetheless, the configuration we employed can be viewed as a stress test, given its resemblance to the common features found in widely utilized laptops and the absence of a high-performance CPU.

\subsection{Experiment 3: Clustering Accuracy Test}

The goal of this final experiment is to address \textbf{RQ3} by testing the utility of the clustering component. It is important to note that the clustering component is not the primary focus of our contribution, so this is not the context to compare our implementation choices with existing alternatives (particularly regarding the embedding techniques and similarity computations used). Instead, the purpose of this test is mainly to assess how effectively this feature assists users in performing an output grouping task that would otherwise require manual effort.


\subsubsection{Using the OntoUML Dataset}
\label{sec:ex3-ontouml}


\noindent\textbf{Data:} As input data, we used the same set of models used in experiment 1. In this experiment, we executed a mining task on the synthetic dataset adopted for experiment 1 and selected 33 out of the whole set of generated patterns, where each selected pattern can be easily traced back to one of the patterns presented in \autoref{ontoUMLpatterns} or one of its variations (e.g., \texttt{Subkind} patterns with a missing \texttt{Subkind}). We then manually clustered all the generated patterns by assigning a cluster label to them (e.g., we labeled the \texttt{Subkind} pattern and its variations as \texttt{cluster\_0}). As the final segmentation, we produced six clusters, each one representing a pattern of \autoref{ontoUMLpatterns}. 

\begin{figure}[h!]
\centering
\includegraphics[width=1\textwidth]{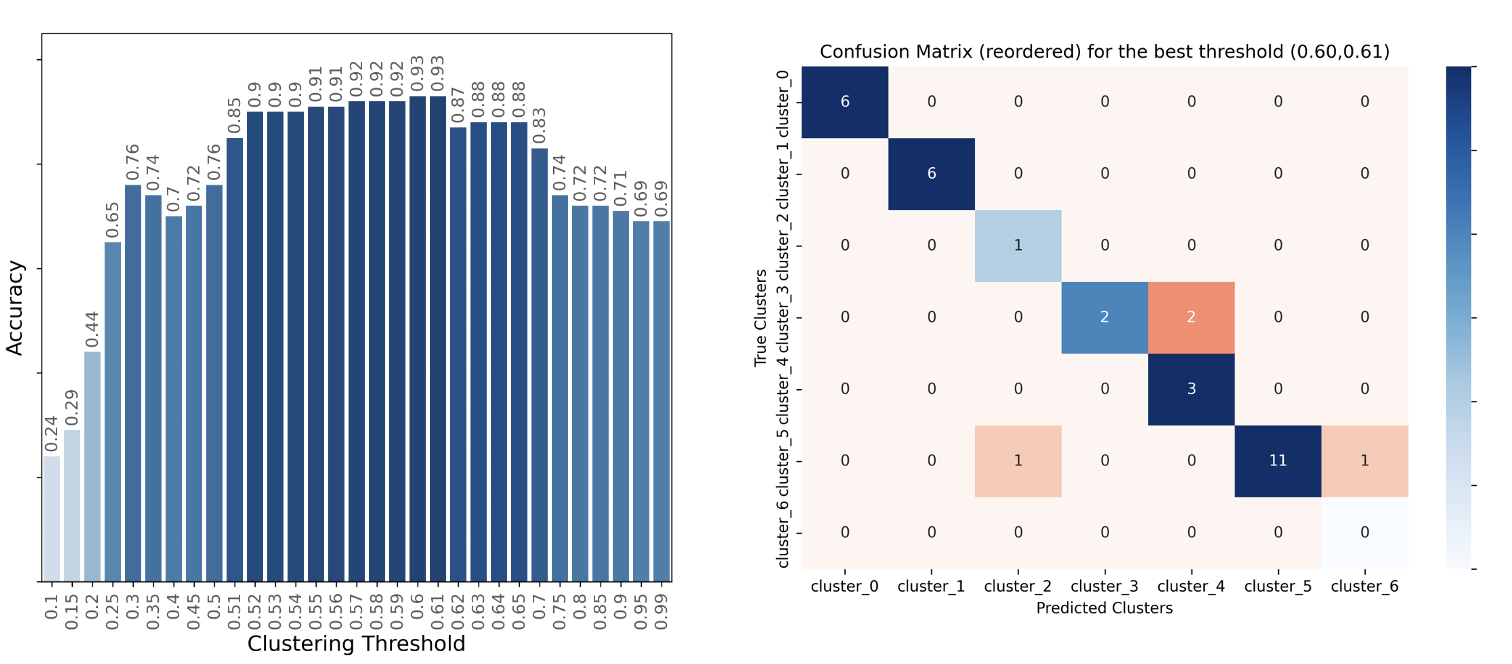}
\caption{Correlation \textit{similarity threshold vs. accuracy (bar plot on the left) and \textit{true clusters vs. predicted clusters ratio} for the best case when there was the number of clusters correspondence (confusion matrix on the right).}}
\label{fig:exp3}
\end{figure}

\vspace{0.5em}
\noindent\textbf{Setup:} We applied our automated clustering step testing multiple similarity threshold values, e.g., 0.1, 0.15, 0.2, 0.25. We then compared the output of each test with the manually created dataset to calculate the accuracy of the automated clustering. Note that, in multiple cases, the number of clusters generated differed from those generated by hand (e.g., when the selected threshold was 0.1, the number of clusters was 2 against 6). For this reason, we calculated the accuracy as the percentage of correctly predicted cluster assignments out of the pairs of patterns being compared. More precisely, the function we adopted calculates the accuracy by comparing pairs of elements on two lists, namely: $P_i$, the list of automatically clustered patterns, and $P_j$, the list of manually clustered patterns. For each pair of patterns $(i, j)$, the function checks whether pattern $i$ and $j$ in both lists $P_i$ and $P_j$ are inserted the same cluster. After checking all pairs $(i, j)$, the function returns the accuracy as the ratio of correctly predicted pairs to the total number of pairs. 

When the accuracy increased significantly, we decreased the distance of the threshold values, to understand what exactly was the point of best performance (for example, instead of going from 0.5 to 0.55, we tested 0.5, 0.51, 0.52, 0.53, 0.54, ..., 0.64, etc.) 
 
\vspace{0.5em}
\noindent\textbf{Results:} We depict the outcomes in \autoref{fig:exp3}. For the dataset we considered, the clustering component we employ in the pipeline achieves 0.93 accuracy when the adopted similarity threshold is 0.6 or 0.61. The increase in accuracy in this is due to the creation of the same number of clusters, such that the different patterns have been distinguished and the variations for each of them not deemed as different patterns. The approach incorrectly predicted only two patterns in the best case, as shown in the confusion matrix in the figure. 

The confusion matrix (right) was reordered using the \textit{Hungarian algorithm},\footnote{\href{https://docs.scipy.org/doc/scipy/reference/generated/scipy.optimize.linear_sum_assignment.html}{\texttt{https://docs.scipy.org/doc/scipy/reference/generated/scipy.optimize.linear\_sum\_assignment.html}}} to optimize alignment between true and predicted clusters, ensuring a clearer visualization of results and accounting for the fact that cluster labels are assigned arbitrarily by the clustering algorithm (e.g., predicted \texttt{cluster\_6} could represent ground truth \texttt{cluster\_1}). Rows represent true clusters (ground truth labels), columns represent predicted clusters and diagonal cells indicate correctly predicted samples, while off-diagonal cells highlight misclassification. The confusion matrix reports the results at the optimal thresholds (0.60–0.61).

\subsubsection{Using the ArchiMate Dataset}

\noindent\textbf{Data:} For this experiment, we used the same dataset of ArchiMate models as in experiment 1 (see \autoref{sec:ex1-archimate}) and we selected 20 patterns from the mining output of the neutral trial in experiment 1. Each selected pattern corresponds to one of the patterns depicted in \autoref{archimatePatterns} or a variation (e.g., missing an element or relationship). The chosen patterns were manually clustered into groups to serve as ground truth in our evaluation (e.g., \texttt{Chatty Service} patterns are assigned the label \texttt{cluster\_0}), resulting in six total clusters, each representing a pattern of \autoref{archimatePatterns}.

\vspace{0.5em}
\noindent\textbf{Setup:} We evaluated the clustering component by applying multiple similarity thresholds (e.g., 0.1, 0.2, ..., 0.9) and comparing the clustering outputs with the ground truth. Initially, to calculate clustering accuracy, we used the same pairwise comparison method as in the OntoUML experiment (see \autoref{sec:ex3-ontouml}). For each pair of patterns, the predicted clustering result was compared with the ground truth to determine if the clustering algorithm accurately grouped or separated them. To refine the evaluation, we further explored thresholds within the interval where the accuracy peaked (e.g., 0.65–0.7) by testing intermediate values (e.g., 0.66, 0.67).

\begin{figure}[!h]
    \centering
    \includegraphics[width=1\linewidth]{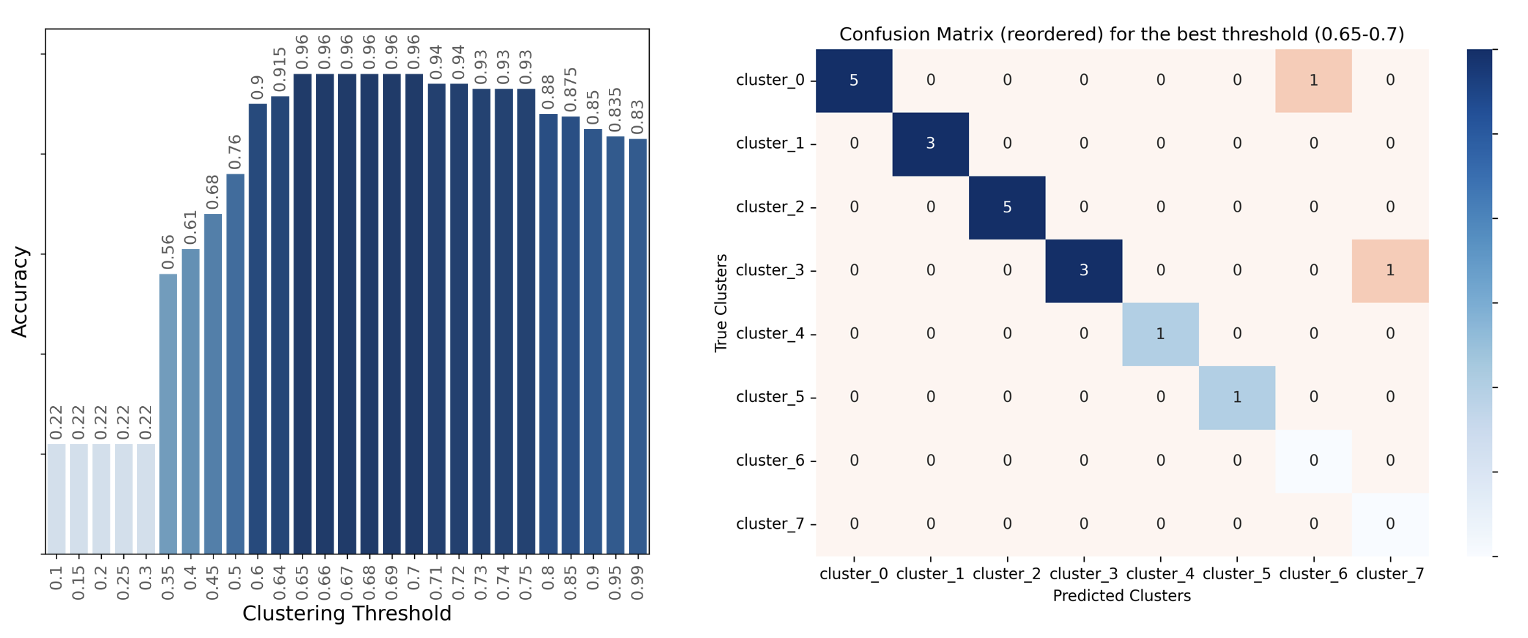}
    \caption{Correlation similarity threshold vs. accuracy (left) and true clusters vs. predicted clusters ratio for the best threshold (right)}
    \label{fig:ex3-archimate}
\end{figure}

\vspace{0.5em}
\noindent\textbf{Results:} The results of this experiment are shown in \autoref{fig:ex3-archimate}. The bar plot (left) shows pair-wise accuracy (from \autoref{sec:ex3-ontouml}) across various clustering thresholds. The highest accuracy, 0.96, was achieved within the range of 0.65–0.7. Beyond this range, accuracy slightly decreased due to over-clustering or merging distinct clusters. The confusion matrix (right) was reordered using the Hungarian algorithm to optimize alignment between true and predicted clusters, as in the OntoUML scenario. The results reported concern the optimal threshold between 0.65 and 0.7.


\subsubsection{Threats to Validity}

A primary concern regarding the validity of this experiment is that we conducted the test over a limited set of patterns by considering one clustering approach only.\footnote{Note that for the current implementation, we adopted a straightforward approach where all vectors must be of the same size, with the same dimensions. Each pattern is represented as a record in a matrix with features such as ``number of nodes'', ``kind'', ``number of relations'', ``general'', and ``specific'', etc. Each feature can hold different values for each pattern (for instance, if a pattern has three nodes classified as ``kind'', it will have ``number of nodes'' = 3 and ``kind'' = 3, while if it lacks nodes for generalizations, it will report 0 for both ``specific'' and ``general'').} Users may vary in their way of classifying patterns, and it could be more challenging when the structures to be classified cannot be straightforwardly traced back to a reference pattern. Still, the test was enough to demonstrate how the clustering approach is reliable in distinguishing the structures that are generated and can be used to organize the data in a manner similar to what the user would adopt.

\section{Application Example}
\label{s:demo}

To illustrate the applicability and utility of our proposed approach, we conducted a simulation of a use case. The primary objective of this simulation was to showcase the effectiveness of our approach in analyzing the practical usage of a particular modeling language: OntoUML. Our specific goal was to demonstrate how the discovery pipeline can be used to extract valuable insights that can enhance our understanding of the language, facilitate modifications, and drive improvements. Within this scope, we have used an extended version of the OntoUML dataset, comprising \textbf{143} models.\footnote{More information about this ever-growing dataset, with statistics about models, can be found in \cite{sales2023fair} and \href{https://github.com/OntoUML/ontouml-models}{\texttt{https://github.com/OntoUML/ontouml-models}}.} To ensure a substantial number of patterns for analysis, we conducted five comprehensive tests, adjusting parameters to avoid generating an overwhelming number of structures.\footnote{Note that the mining process should be viewed
as an iterative procedure, concluding once interesting results are obtained, so it becomes difficult to envision an optimal set of parameter \textit{a priori}. Testing the current approach to establish optimal mining parameters and exploit the pipeline to generate a robust dataset of patterns could be a valuable objective for future research.} In a second step, we went through the whole set of generated patterns and we engaged in internal discussions with two experts who contributed to the design of the language.\footnote{Note that we have deliberately separated this section from the experiments section. The reason is that in this case we did not properly design an experiment, but simulated a use case with two language designers, who also participated in the requirements definition and design of the approach we are offering. That's why we talked about ``demonstration'' and ``internal'' discussion.} The interaction with the experts occurred in two stages. First, we presented the generated patterns and asked them to identify the patterns for which it was worth having a more in-depth discussion. Second, after defining a subset of 40 patterns, we went through a discussion of each of them,\footnote{The examples in this section can be viewed as a summary of discussions on the most relevant patterns. We added this clarification as a note at the location indicated by the reviewer.} focusing primarily on 
\begin{enumerate*}[label=\textit{.\roman*}]
\item the unexpectedness of the mined structure, and 
\item the possibility of reusing the structure somehow. 
\end{enumerate*}
This simulation allowed us to address a new research question: 
\vspace{0.5em}
\begin{itemize}
 \setlength{\itemindent}{1.0em}
    \item [\textbf{RQ4}] \textit{To what extent can the approach be used to discover new interesting structures?} This research question relates to \textbf{R1}, \textbf{R2} and \textbf{R3}, and explores the approach's capability to identify compelling structures—potentially unexpected ones—in real-world scenarios featuring a diverse set of models with varying levels of complexity. If identified, these structures can offer valuable insights for language engineers, potentially informing their future design strategies. 
\end{itemize}
\vspace{0.5em}
Next, we present four exemplary structures, taken from the subset of selected interesting patterns, which represent significant findings from our demonstration. Each example is accompanied by a brief discussion outlining the actions that could be undertaken based on our observations.

\vspace{0.5em}
\noindent\textbf{Example 1:} In OntoUML, \texttt{Mode}s are concepts representing particular types of properties with no structured values, which depend on their bearers \cite{guizzardi2021types}. According to the OntoUML specifications provided in \cite{guizzardi2005ontological} the constraints for modes are the following: 

\begin{enumerate}[label=\textit{.\roman*}]
\item every \texttt{Mode} must be (directly or indirectly) connected to an association end of at least one \texttt{Characterization} relation, and
\item the multiplicity of the characterized end (opposite to the \texttt{Mode}) must be exactly one. 
\end{enumerate}

Therefore, the situation in which a \texttt{Mode} is connected to two different bearers is not admitted. 

\begin{figure}[h!]
\centering
\includegraphics[width=1\textwidth]{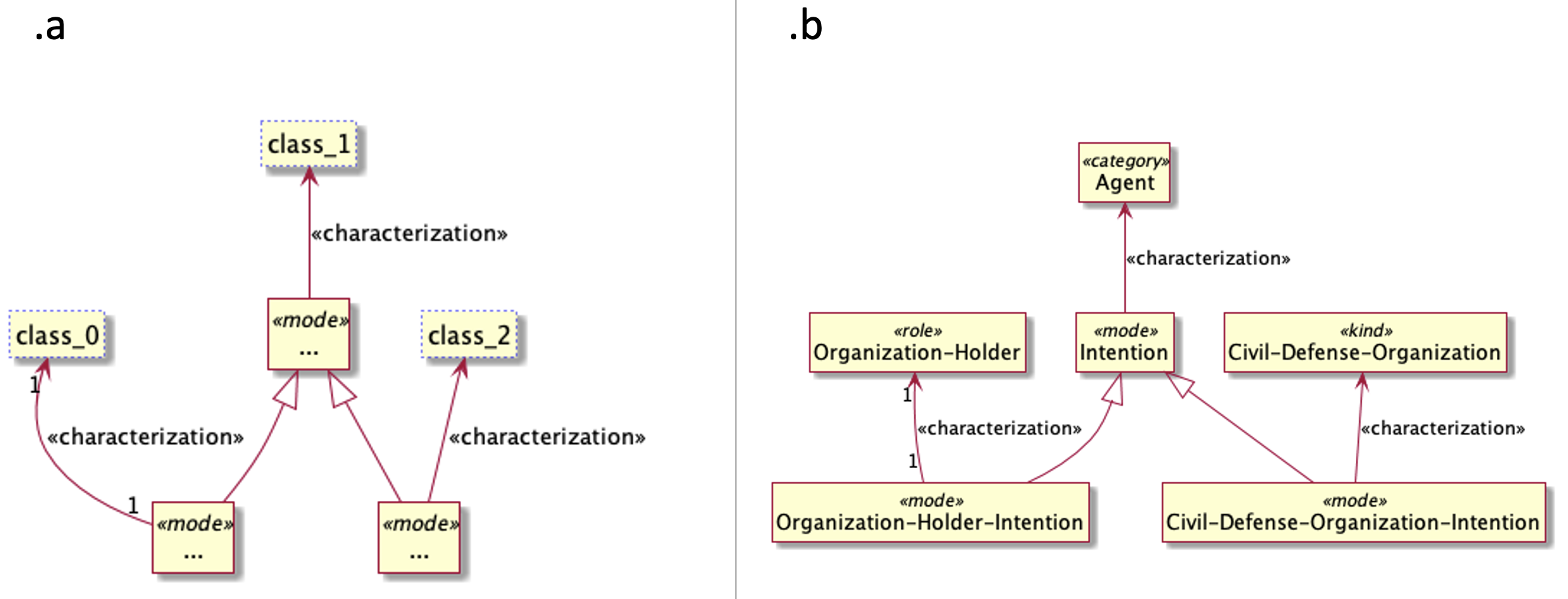}
\caption{Interesting structure: \texttt{Mode}s \textit{typing and multiple} \texttt{Characterization}s. \textit{.a} found structure; \textit{.b} example occurrence.}
\label{fig:modes}
\end{figure}

\autoref{fig:modes} shows a structure with three \texttt{Mode}s, each one of them characterizing a distinct bearer, where two \texttt{Mode}s are subtypes of a parent \texttt{Mode}.\footnote{Nodes with dashed line denoted as \texttt{class\_0} or \texttt{class\_1}, i.e., without associated stereotype, are nodes that in the discovered patterns present relationships about which we have information related to the source node, but not the target node, or vice versa. This is because, as mentioned above, the graphs given as input to the mining algorithm have all relationships reified. So, for example, in \autoref{fig:modes} \texttt{class\_1} was created because in the pattern we had a \texttt{mode} node associated with an edge \textit{source} to a relation \textit{characterization}.}

The information related to the occurrence of the pattern (\textit{.b}) reveals that the reference bearers are represented by a \texttt{Role}, a \texttt{Category}, and a \texttt{Kind}, respectively. This does not yet present an actual problem; however, it may involve the generation of scenarios that are not intended. In fact, the pattern suggests checking whether \texttt{role} and \texttt{kind} are subtypes of the same concept. In fact, if this is not the case, the child \texttt{mode}s might end up associated with two different \texttt{bearers} (e.g., \texttt{class\_0} and \texttt{class\_1} may classify different instances). This specific situation may be avoided by forcing the bearers of the child \texttt{mode}s (e.g., \texttt{class\_0} and \texttt{class\_2}) to be subclasses of the bearer of the main parent \texttt{mode}. In conclusion, the recurring structure we found does not present an incorrect practice, but it may trigger an evolution of the language. Originally, in fact, the designers of OntoUML did not dwell on the possibility of constructing taxonomies of \texttt{mode}s. However, through the results of this mining activity, emerges how the \textit{de facto} modelers use these taxonomies (the importance of which was also discussed in \cite{guizzardi2018endurant}). This uncovered structure, then, opens the door to introducing new possible language constraints to avoid unwanted configurations.

\begin{figure}[h!]
\centering
\includegraphics[width=0.9\textwidth]{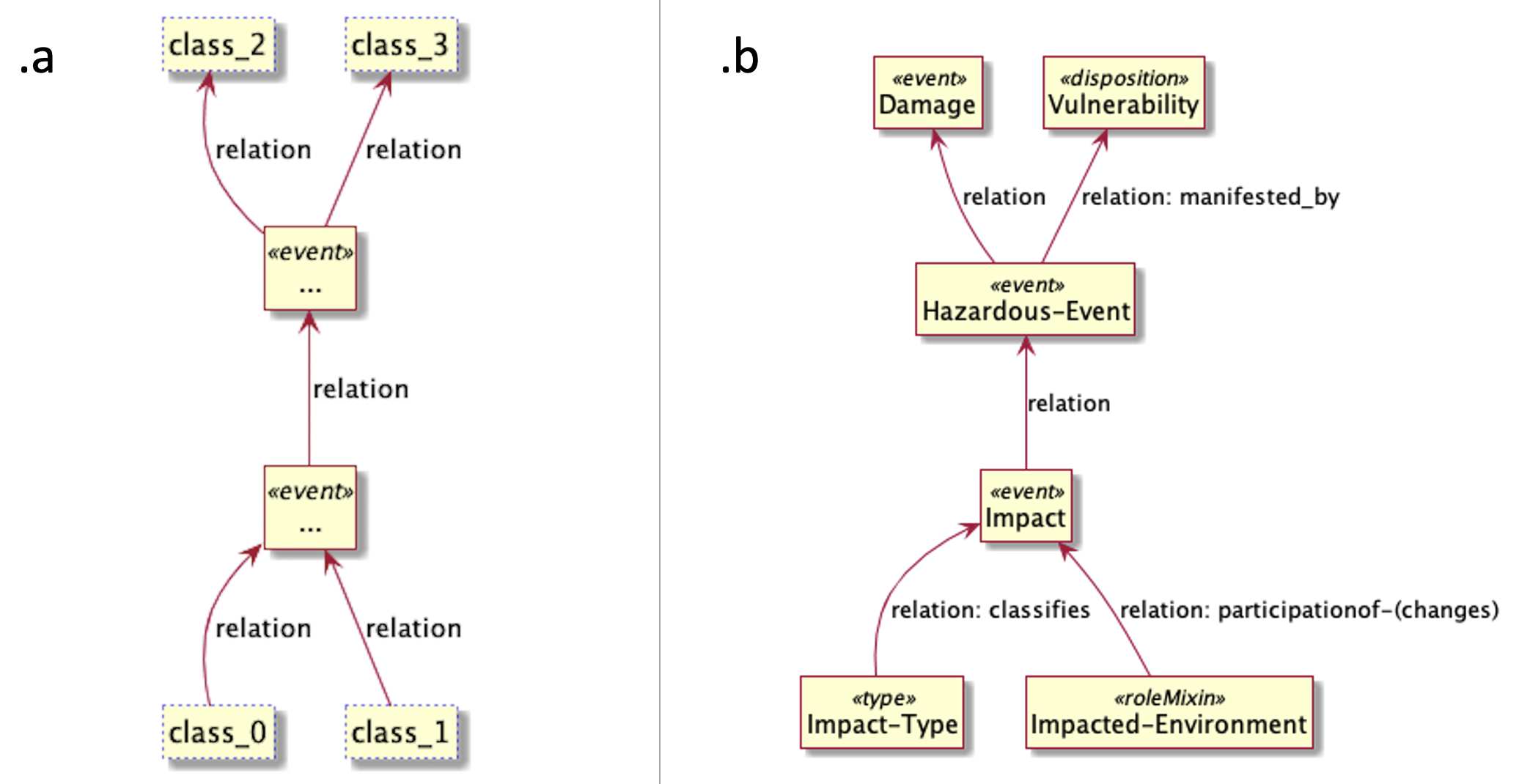}
\caption{Interesting structure: \texttt{Event}s \textit{and their relations}. \textit{.a} found structure; \textit{.b} example identified.}
\label{fig:events}
\end{figure}

\vspace{0.5em}
\noindent\textbf{Example 2:} The structure provided by \autoref{fig:events} is significant for understanding how \texttt{Event}s and their relationships are used in OntoUML. 

Besides the structure we report here, many other variations having similar characteristics have been found. Consequently, a useful lesson we can learn from this is that the OntoUML module concerning events is less in control by modelers than that concerning, for example, \texttt{Kind}s or \texttt{Role}s. OntoUML has several relationships that can be used between \texttt{Event}s. For example, see the common participation relationship or the \texttt{HistoricalDependence} relationships. The fact that the structure in \autoref{fig:events} and many similar ones do not have stereotypes for relations between \texttt{Event}s can certainly be a sign that the users of the language do not properly understand these. In addition, by looking at the pattern occurrence information, one can see that some users have used relationship labels that invoke a specific stereotype (e.g., the case of \texttt{ParticipationOf} in the figure). In conclusion, this identified structure emphasizes the need for development guidelines for any related patterns for \texttt{Event} modeling.

\begin{figure}[h!]
\centering
\includegraphics[width=1\textwidth]{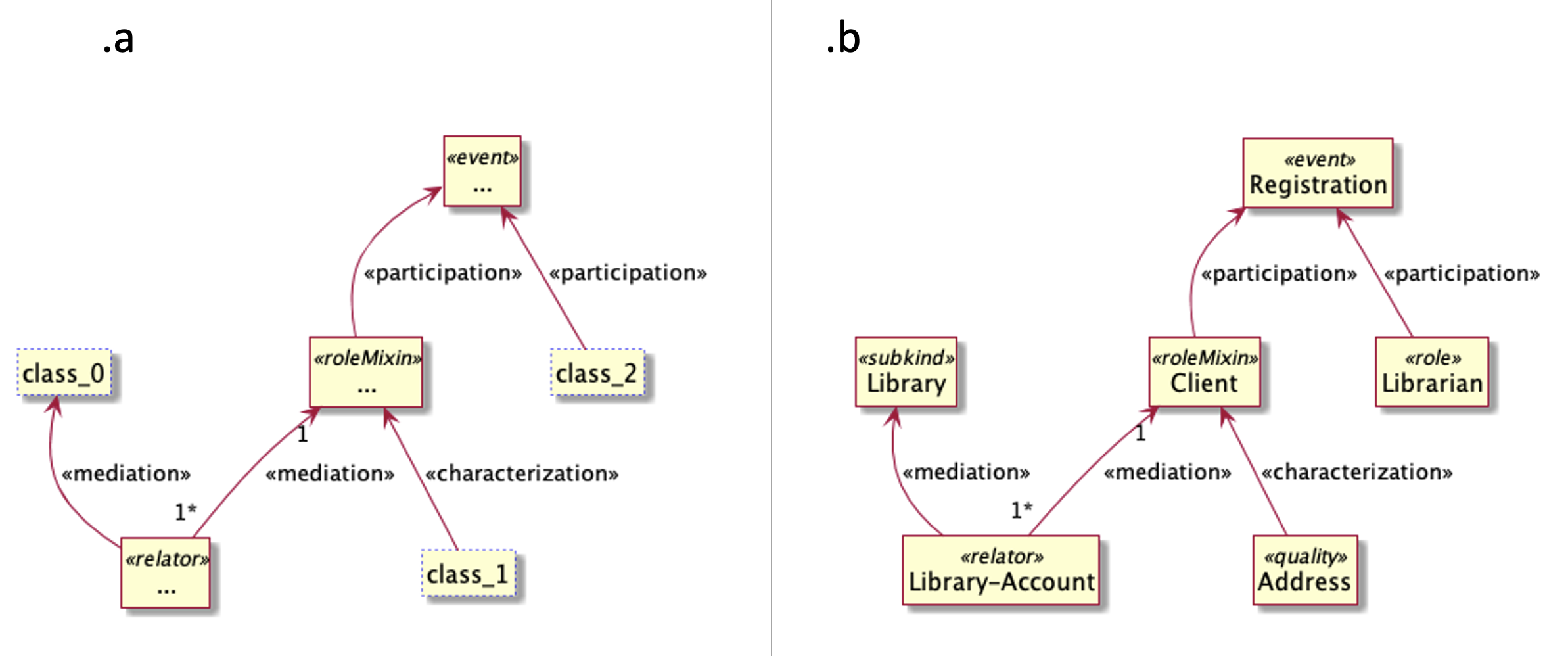}
\caption{Interesting structure: \texttt{Relator}, \texttt{RoleMixin} \textit{and} \texttt{Event} \textit{structure}. \textit{.a} found structure; \textit{.b} example occurrence.}
\label{fig:events-rMixin}
\end{figure}

\vspace{0.5em}
\noindent\textbf{Example 3:} This example's structure still relates to \texttt{Event} modeling. Here, what we represent in \autoref{fig:events-rMixin} could be an occurrence of a syntactical mistake, depending on the cardinalities of the participation relation between event and \texttt{RoleMixin}. Here, the concept of \texttt{RoleMixin} \cite{guizzardi2005ontological} is the equivalent of \texttt{Role}, but instead of being played by instances of the same \texttt{Kind} (e.g., student-person), it is played by instances of different \texttt{Kind}s. Examples of \texttt{RoleMixin}s include customers and insured items, the former being played by both people and companies, whilst the latter being played by cars, houses, and paintings \cite{sales2015ontological}.

That being considered, the OntoUML \texttt{RoleMixin} construct requires that there are at least two identity providers involved and then at least two different \texttt{Role} instances. Thus, in what sense does a \texttt{RoleMixin} participate in an \texttt{Event}? Looking at the occurrence of the pattern, the concept of \texttt{RoleMixin} is modeled as ``client'' and in this sense, we may have different instances of this class that are, for instance, persons or organizations. Considering this structure, to avoid possible mistakes, the cardinalities of the relationship between the \texttt{RoleMixin} concept and the \texttt{Event} concept must be made explicit. The recurrent structure we found indicates that the modeler left this information implicit. Consequently, the possible insight that can be collected by the language engineer through the assessment of this structure is that the modeler should be guided in making explicit the connection between the identity providers and the given \texttt{Event}, possibly by providing the correct cardinalities of the associations in question.

\section{Discussion}
\label{s:discussion}
\vspace{0.5em}
\noindent\textbf{RQ1: Can the proposed approach generate structures encoding previously recognized interesting patterns?} The answer to this question is affirmative. Experiment 1 demonstrated that already-known structures can be discovered accounting for all the constructs of which they are composed and without missing any occurrence. For the patterns we selected, we successfully discovered all classes of patterns and their occurrences. The patterns we have chosen were also useful to demonstrate how the approach can account for the core constructs in the input language. The experiment also highlighted the importance of customization options is important in reducing the number of results, thus facilitating the identification process. The importance is enlarged when the methodology is applied to extensive real-world datasets, such as the one represented by the OntoUML catalog.

\vspace{0.5em}
\noindent\textbf{RQ2: Which are the main parameters affecting the performance of the proposed approach?} According to \autoref{perf} and \autoref{ex2output-archimate}, the increase in pattern size and minimum support value improves the average performance but decreases the number of patterns that can be found. Unlike when the goal is to search for more frequent structures, we can conclude from this that when it is necessary to find more information or possibly unexpected information, it is necessary to find the right trade-off between effectiveness and performance. On a dataset comprising OntoUML or ArchiMate models, if we search too small and too infrequent patterns, the algorithm can produce even more than 10.000 outputs, often very similar to each other. Here, too, the role of customization steps can be essential, especially in reducing the information in the graphs to be sent as input to the mining algorithm and in eliminating redundant or unwanted outputs, thus also reducing the time of visualization and assessment.

\vspace{0.5em}
\noindent\textbf{RQ3: Is the clustering step accurate in grouping structures?} Thanks to Experiment 3, we observe how our proposed clustering component provides a grouping criterion for patterns akin to manual user-performed ones. Our current configuration enables the differentiation of various structures by considering both the node count and the types of constructs involved. This capability positions it as a valuable asset during the results assessment phase. However, a point of discussion centers on the heterogeneity of the threshold values needed to achieve a satisfactory grouping, which, as we have found in other tests, can vary from case to case. We cannot assert a universally optimal threshold, such as adhering to a specific value like 5, for attaining the best grouping. Our trials have shown that optimal thresholds may range around 0.6 or 0.7 
in some instances. One consistent observation, though, is that the optimal range for classification consistently falls within the mid-range of the scale. That is, thresholds between 0.1 and 0.3 or between 0.8 and 1.0 do not consistently achieve the most effective clustering. Considering the iterative nature of the discovery process, where users often experiment and fine-tune parameters to get the desired output in terms of size and type of structures, adjusting threshold values to achieve the desired clustering is also part of the iterative discovery process.

\vspace{0.5em}
\noindent\textbf{RQ4: To what extent is the approach useful for discovering new interesting structures?} From the tests we did for the utility demonstration, we observed that to find unexpected modeling practices, the importance value of frequency (or minimum support) decreases. In fact, if parameters are set to find frequent structures, it is more likely that we will find already known structures, especially with a language like OntoUML, where modelers operate following predefined guidelines and patterns. This mainly has a negative effect on the overall performance of the approach. In fact, as we saw in Experiment 2, the more we decrease the frequency threshold, the longer the mining process takes. Moreover, the approach has the potential to generate patterns of a very large order of magnitude. However, this did not prevent us from collecting useful and interesting information, such as that discussed in the previous section, which can trigger a range of analyses and possibly interventions by the language engineers that they had not thought of before. Currently, our emphasis is not on exploring the interesting types and unexpected structures that can be found, as it is not the main purpose of the paper. Our understanding of classifying these outputs is growing as we consider the analyzes and operations they might trigger. 

\textit{Structures triggering the definition of new patterns or anti-pattern}. For instance, structures like the one represented in \autoref{fig:modes} could be a good start for understanding and, possibly, devising new modeling strategies to be suggested to modelers. In the specific case of the example given, the structure of modes is a concrete case for understanding how to enable the engineers to avoid generating unintended instantiations. For example, engineers can concretely achieve this by exploiting the found structure and offering an ideal way of modeling \texttt{mode}s taxonomies, or by introducing new constraints that force the creation of generalization links between the concepts characterized by the sub-\texttt{mode}s and the concept characterized by the parent-\texttt{mode}. 

\textit{Structures triggering a clarification of constructs and their usages}. A typical example in this regard is that represented by \autoref{fig:events}. Here, the pervasive absence of relation stereotypes composing structures involving events is a clear sign of how the modeling guidelines related to the area of the language devoted to these constructs (in this case, the area dependent on UFO-B~\cite{guizzardi2013towards}) can be further elaborated.

\textit{Structures that highlight possible adoptions of bad practices}. In this sense, the approach is useful in unearthing new types of errors or understanding whether practices deemed to be avoided are actually adopted, and to what extent. The example provided by \autoref{fig:events-rMixin} is a typical case and demonstrates the utility of the approach since it can be a strong ally in understanding how to possibly guide modelers to not omit key information or extend the language with new constraints.

\section{Related Work}
\label{s:relatedwork}


There is extensive literature~\cite{Fournier2017survey,gan2019survey,Fournier2022challenges,guvenoglu2018qualitative} on pattern discovery and its applications in a variety of domains, including software code~\cite{INTiMALS-DS2019}, databases~\cite{Fournier2014SPMF}, educational processes~\cite{Bogarin2018edu}, business processes~\cite{tiwari2008review}, model-driven engineering (MDE)~\cite{pescador2015pattern}, EMF metamodels \cite{babur2024language}, etc. 
However, to the best of our knowledge, there is a significant potential for research into automatic applications that support pattern discovery in conceptual models.

In this focused area of research, the closest work to what we propose is that of Skouradaki et al. \cite{skouradaki2016rose}, who designed a pattern mining algorithm for \textit{BPMN}. Still, the goal of our contribution is not to provide a new mining algorithm. Our emphasis predominantly lies in the combination of a well-established FSM technique with graph manipulation techniques. Furthermore, a considerable amount of effort from our side concerns the definition of an interactive process where users can participate in the discovery activities, thus affecting the reliability of the final output. Last, we designed the approach with the scope of covering different conceptual modeling languages by keeping all the functions of the approach as language-independent. 

{\L}awrynowicz et al. \cite{lawrynowicz2018discovery} seek to discover domain patterns related to specific areas of information and independent of the modeling language constructs that recur across \textit{OWL ontologies} by applying a \textit{tree-mining} technique. 
They divide their contribution into two main steps, which partially resemble aspects of our strategy, namely: a transformation step, where ontology axioms are transformed into tree structures; and an association analysis step, where co-occurring axioms are extracted to discover ontology patterns. This research is applied to a set of ontologies from the \textit{BioPortal} repository and is very similar to ours in spirit. However, our solution presents key differences. 
First, for the mining step, we adopted the \textit{frequent subgraph mining} algorithm, thus involving a completely different input preparation step. Second, we devised our approach with the main goal of discovering \textit{structural modeling patterns}, namely patterns defined simply by the combination of constructs of a modeling language. In {\L}awrynowicz et al.'s work, the discovered patterns concern primarily domain-specific information that may recur within or across ontologies (e.g., what are the recurrent properties of the class ``person'')~\cite{lawrynowicz2018discovery}. Again, the interaction capabilities we proposed are out of their scope.

In the same direction, Lee et al. \cite{lee2021learning} seek to discover domain patterns across and within ontologies. However, to address this challenge, two different steps are adopted: a step where sub-graphs are extracted through candidate generation and chunking processes; a step where \textit{frequent sub-graphs mining} \cite{ramraj2015frequent} is adopted. This work also focuses on domain-specific patterns and one of its priorities is to allow the processing of large-scale knowledge graphs. Furthermore, the paper lacks instructions on how to handle an interactive discovery process.

Two other approaches that are worth mentioning are the ones presented by Ardimento et al.~\cite{mod2023pattern} and the one that was introduced by Fumagalli et al.~\cite{fumagalli2022pattern}. Both refer directly to UML class diagrams as the reference modeling language to be mined. However, the former is more aimed at mining modeling events that occur on \textit{Visual Paradigm},\footnote{\href{https://www.visual-paradigm.com/download/community.jsp}{\texttt{https://www.visual-paradigm.com/download/community.jsp}}} i.e., the reference editing tool, and not properly frequent subgraph structures. The second, on the other hand, is much closer to our approach and is primarily focused on using frequent item set mining, instead of frequent subgraph mining, and also does not introduce the set of facilities we have offered here to support the discovery process.

Along similar lines, particularly within the context of MDE and, more broadly, software engineering, other solutions exist that address problems similar to ours.

For instance, the work in~\cite{kaczor2010identification} aims to enhance software maintenance by identifying design motifs, which are solutions to recurring design challenges. The authors employ two pattern-matching algorithms adapted from bioinformatics—automata simulation and bit-vector processing—to detect exact and approximate occurrences of design motifs in object-oriented code. The findings from this research underscore the potential of bioinformatics-inspired graph-matching methods to facilitate design recovery, streamline the analysis of complex software architectures, and uncover embedded design patterns.

Likewise, the study in~\cite{gueheneuc2010improving} tackles the issues of recognizing design patterns in object-oriented programs. The proposed approach utilizes design motifs to identify pattern instances and filter out irrelevant matches. This strategy effectively narrows the search space, aiding the pattern discovery process.

To summarize, our approach distinguishes itself from related work through two key novel contributions. First, we empower users in their discovery process by offering a comprehensive set of interactive steps, providing guidance and assistance throughout the discovery journey. Second, we employ frequent subgraph mining, a robust established mining technique, for mining subgraphs in diagrams representing models in different languages. 
Our approach also accounts for all vital elements of these diagrams, including cardinalities, multiple edge labels and class labels, and multi-directionality, thus ensuring the possibility of finding patterns containing all this information.

As a final remark, the approach we propose employs various techniques across each module in the pipeline. For example, we use a technique for mining frequent structures, a technique for matching to trace domain information, a technique for transforming mining algorithm outputs into vectors, and a technique for clustering these vectors. Each of these techniques could merit its own discussion of related work, spanning fields such as ``metamodel clone detection'', ``linguistic corpus analysis'', ``efficient pattern similarity computation and clustering'', and ``plagiarism detection''.

However, we have focused our related work review on approaches that address our specific problem: the automatic support for identifying recurrent structure heuristics. Therefore, where relevant, we have included information on complementary techniques related to individual pipeline steps within the sections dedicated to implementation choices.

\section{Final Considerations}
\label{s:considerations}


This paper presents a practical interactive approach, whose implementation is available at \href{https://github.com/unibz-core/CM-Mining}{\texttt{https://github.com/unibz-core/CM-Mining}}, for automating the empirical discovery of recurrent structures in conceptual models by combining state-of-the-art graph manipulation techniques and frequent subgraph mining. Structural patterns can be exploited to identify reusable representations of bad or good modeling practices. They are typically harvested by manual labor-intensive processes, which usually take a long time to converge into pattern catalogs. Our automated process for pattern discovery facilitates the construction of pattern catalogs tailored for modeling languages. Additionally, we create a mechanism for helping language engineers to create higher-granularity primitives in that language, i.e., modeling patterns that can become part of the grammar and tools of that language~\cite{guizzardi2014ontological}.

Based on the encouraging results presented here, we created a list of tasks for immediate future work. First, we are going to further assess the approach to finding unexpected patterns.  In order to address this, we will involve representative users and run an evaluation through usability testing. Second, we are going to test the approach with models encoded in new conceptual modeling languages, such as BPMN (\textbf{R5}). 

We chose to validate the proposed approach using two state-of-the-art conceptual modeling languages, OntoUML and ArchiMate. However, our solution is designed around graph-structured data, allowing it to encode any conceptual model constructs. Furthermore, the approach is fully language-independent, apart from the initial importing and the final visualization steps, and can be adapted to various input formats.

Our focus on these two languages and their associated datasets was driven by several key factors. First, both languages include constructs—such as classes, annotations, and cardinalities—that are also prevalent in widely-used languages like UML and BPMN. Additionally, these languages are accompanied by two high-quality, scientifically validated datasets that adhere to FAIR principles. Finally, we had direct access to the language designers, who provided essential support and feedback on the functionalities and output structures required throughout our research.

\backmatter




\bibliography{main}

\end{document}